\documentclass[10pt,journal,compsoc]{IEEEtran}
\ifCLASSOPTIONcompsoc
  \usepackage[compress]{cite}
\else
  \usepackage{cite}
  \fi
\usepackage{cite}
\usepackage[pdftex]{graphicx}
\usepackage{caption}
\usepackage{subcaption}
\usepackage{amsmath}
\usepackage{algorithmic}
\usepackage{array}
\usepackage{import}
\usepackage{amssymb}
\usepackage{times}
\usepackage{epsfig}
\usepackage{graphicx}
\usepackage{commath}
\usepackage[colorinlistoftodos]{todonotes}
\usepackage{stfloats}
\usepackage{url}
\usepackage[colorlinks]{hyperref}
\usepackage[author={Max Schlepzig}]{pdfcomment}
\usepackage{multirow}
\usepackage{xspace}
\usepackage{hyperref}
\hypersetup{citecolor=blue}
\usepackage{verbatim}
\usepackage[export]{adjustbox}

\makeatletter
\def\footnoterule{\kern-3\p@
  \hrule \@width 2in \kern 2.6\p@} % the \hrule is .4pt high
\makeatother

\newcommand{\csgnet}{\textsc{CSGNet}\xspace}
\newcommand{\scsgnet}{\textsc{CSGNetStack}\xspace}
\newcommand{\eg}{\textit{e.g.}\xspace}
\newcommand{\etal}{\textit{et al.}\xspace}
\newcommand{\ie}{\textit{i}.\textit{e}.\xspace}
\newcommand{\etc}{\textit{etc}.\xspace}
\begin{document}

\title{Neural Shape Parsers for \\
Constructive Solid Geometry}

% \author{Michael~Shell,~\IEEEmembership{Member,~IEEE,}
%         John~Doe,~\IEEEmembership{Fellow,~OSA,}
%         and~Jane~Doe,~\IEEEmembership{Life~Fellow,~IEEE}% <-this % stops a space
% \IEEEcompsocitemizethanks{\IEEEcompsocthanksitem M. Shell was with the Department
% of Electrical and Computer Engineering, Georgia Institute of Technology, Atlanta,
% GA, 30332.\protect\\
% E-mail: see http://www.michaelshell.org/contact.html
% \IEEEcompsocthanksitem J. Doe and J. Doe are with Anonymous University.}% <-this % stops an unwanted space
% \thanks{Manuscript received April 19, 2005; revised August 26, 2015.}}
\author{$\text{Gopal Sharma}^{\dagger}$ \quad  $\text{Rishabh Goyal}^{\daleth}$ \quad $\text{Difan Liu}^{\dagger}$ \quad $\text{Evangelos Kalogerakis}^{\dagger}$ \quad $\text{Subhransu Maji}^{\dagger}$ \\
University of Massachusetts, Amherst$^{\dagger}$, University of Illinois at Urbana-Champaign$^\daleth$\\
{\tt\small \{gopalsharma,dliu,kalo,smaji\}@cs.umass.edu$^{\dagger}$, rishgoyell@gmail.com$^{\daleth}$}\\
}
% \author{Gopal~Sharma \quad Subhransu~Maji \quad Evangelos~Kalogerakis}
%\markboth{Journal of \LaTeX\ Class Files,~Vol.~14, No.~8, August~2015}%
%{Shell \MakeLowercase{\textit{et al.}}: Bare Demo of IEEEtran.cls for Computer Society Journals}
% *** Note that you probably will NOT want to include the author's ***
% *** name in the headers of peer review papers.                   ***
% You can use \ifCLASSOPTIONpeerreview for conditional compilation here if
% you desire.
%\IEEEspecialpapernotice{(Invited Paper)

\IEEEtitleabstractindextext{
\begin{abstract}
  Constructive Solid Geometry (CSG) is a geometric modeling technique that
  defines complex shapes by recursively applying boolean operations on
  primitives such as spheres and cylinders. We present \csgnet, a deep network
  architecture that takes as input a 2D or 3D shape and outputs a CSG program
  that models it. Parsing shapes into CSG programs is desirable as it yields a
  compact and interpretable generative model. However, the task is challenging
  since the space of primitives and their combinations can be prohibitively
  large. \csgnet uses a convolutional encoder and recurrent decoder based on
  deep networks to map shapes to modeling instructions in a feed-forward manner
  and is significantly faster than bottom-up approaches. We investigate two
  architectures for this task --- a vanilla encoder (CNN) - decoder (RNN) and
  another architecture that augments the encoder with an explicit memory module
  based on the program execution stack. The stack augmentation improves the
  reconstruction quality of the generated shape and learning efficiency. Our
  approach is also more effective as a shape primitive detector compared to a
  state-of-the-art object detector. Finally, we demonstrate \csgnet can be
  trained on novel datasets without program annotations through policy gradient
  techniques.
\end{abstract}

% Note that keywords are not normally used for peerreview papers.
\begin{IEEEkeywords}
Constructive Solid Geometry, Reinforcement Learning, Shape Parsing.
\end{IEEEkeywords}}

% % make the title area
% %\listoftodos
\maketitle

\IEEEpeerreviewmaketitle
% -----------------------------------------------------------------------
\IEEEraisesectionheading{\section{Introduction}\label{sec:introduction}} 
In
recent years, there has been a growing interest in generative models of 2D or 3D
shapes, especially through the use of deep neural networks as image or shape
priors. %~\cite{kulkarni2015deep,Choy,Dosovitskiy2017,FanSG17}.
However, current methods are limited to the generation of low-level shape
representations consisting of pixels, voxels, or points. Human designers, on the
other hand, rarely model shapes as a collection of these individual elements.
For example, in vector graphics modeling packages (\eg, Inkscape, Illustrator,
and so on), shapes are often created through higher-level primitives, such as
parametric curves (\eg, Bezier curves) or basic shapes (\eg, circles,
polygons), as well as operations acting on these primitives, such as boolean
operations, deformations, extrusions, and so on. Describing shapes with
higher-level primitives and operations is highly desirable for designers since
it is compact and makes subsequent editing easier. It may also better capture
certain aspects of human shape perception such as view invariance,
compositionality, and symmetry~\cite{biederman1987recognition}.

\begin{figure}[!htbp] \centering
  \includegraphics[width=0.9\linewidth]{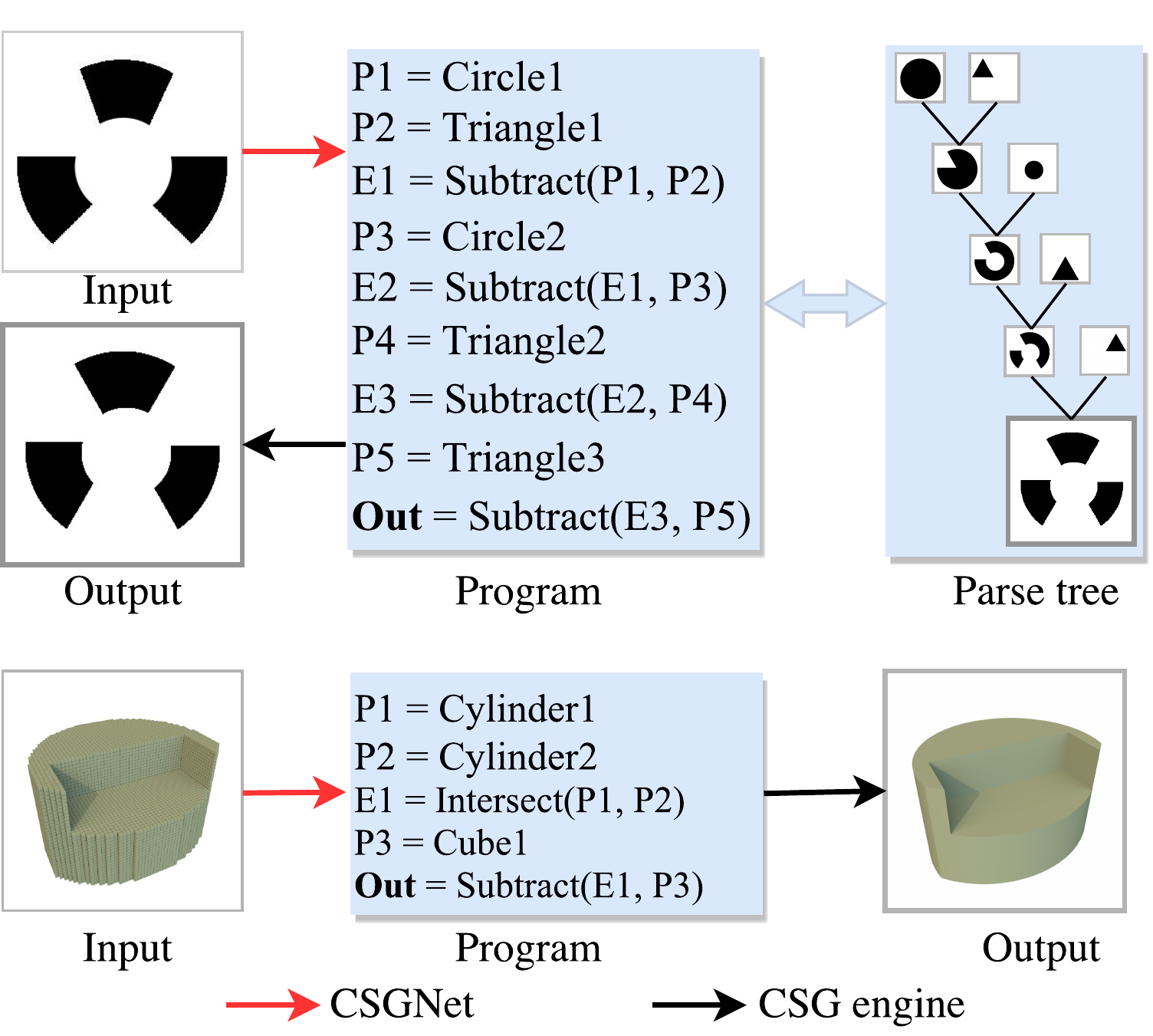} \caption{\label{fig:teaser}\textbf{Our
      shape parser produces a compact program that generates an input 2D\ or 3D\
      shape}. On top is an input image of 2D\ shape, its program and the
    underlying parse tree where primitives are combined with boolean operations.
    On the bottom is an input voxelized 3D\ shape, the induced program, and the
    resulting shape from its execution.}
  \vskip -5mm \end{figure}

The goal of our work is to develop an algorithm that parses shapes into their
constituent modeling primitives and operations within the framework of
Constructive Solid Geometry (CSG)~\cite{Laidlaw1986CSG}. CSG is a popular
geometric modeling framework where shapes are generated by recursively applying
boolean operations, such as union or intersection, on simple geometric
primitives, such as spheres or cylinders. Figure~\ref{fig:teaser} illustrates an
example where a 2D shape (top) and a 3D shape (bottom) are generated as a
sequence of operations over primitives or a \emph{visual program}. Yet, parsing
a shape into its CSG program poses a number of challenges. First, the number of
primitives and operations is not the same for all shapes \ie, our output does
not have constant dimensionality, as in the case of pixel arrays, voxel grids,
or fixed point sets. Second, the order of these instructions matter --- small
changes in the order of operations can significantly change the generated shape.
Third, the number of possible programs grows exponentially with program length.

Existing approaches for CSG parsing are predominantly search-based. A
significant portion of related literature has focused on approaches to
efficiently estimate primitives in a bottom-up manner, and to search for their
combinations using heuristic optimization. While these techniques can generate
complex shapes, they are prone to noise in the input and are generally slow. Our
contribution is a neural network architecture called \csgnet that generates the
program in a feed-forward manner. The approach is inspired by the ability of
deep networks for generative sequence modeling such as for speech and language.
As a result \csgnet is efficient at test time, as it can be viewed as an
\emph{amortized search} \cite{Gershman:2014wt} procedure. Furthermore, it be used as an initialization
for search-based
approaches leading to improvements in accuracy at the cost of computation.

At a high-level, \csgnet is an encoder-decoder architecture that encodes the
input shape using a convolutional network and decodes it into a sequence of
instructions using a recurrent network (Figure~\ref{fig:overview}). It
is trained on a large synthetic dataset of automatically generated 2D and 3D
programs (Table~\ref{table:dataset}). However, this leads to poor generalization
when applied to new domains. To adapt models to new domains without program
annotations, we employ policy gradient techniques from the reinforcement learning
literature~\cite{Williams92simplestatistical}. Combining the parser with a CSG
rendering engine allows the networks to receive feedback based on the visual
difference between the input and generated shape, and the parser is trained to
minimize this difference (Figure~\ref{fig:overview}). Furthermore, we
investigate two network architectures: a vanilla recurrent network (\csgnet),
and a new variant (\scsgnet) that incorporates explicit memory in a  form of a
stack, as seen in Figure~\ref{fig:arch}. The memory is based on the execution
stack of the CSG engine and enables explicit modeling of the intermediate
program state. Our experiments demonstrate that this improves the overall
accuracy of the generated programs while using less training data.

We evaluate the \csgnet and \scsgnet architectures on a number of shape parsing
tasks. Both offer consistently better performance than a nearest-neighbor
baseline and are significantly more efficient than an optimization based
approach. Reinforcement learning improves their performance when applying them
to new domains without requiring ground-truth program annotations making the
approach more practical (Table~\ref{table:CAD}). We also investigate the role of
size of training data and reward shaping on the performance of the parser.
Finally, we evaluate the performance on the task of primitive detection and
compare it with a Faster R-CNN detector~\cite{NIPS2015-5638} trained on the same
dataset. \csgnet offers $4.2\%$ higher Mean Average Precision (MAP) and is $4$
times faster compared to the Faster R-CNN detector, suggesting that joint
reasoning about the presence and ordering of objects leads to better performance
for object detection (Table~\ref{tab:primitive-detection}).

This paper extends our work that first appeared in~\cite{Sharma2018}, adding
analysis on reward shaping and the dependence on training set size, as well as
the stack-augmented network architecture. Our PyTorch~\cite{pytorch}
implementation is publicly available at:
\url{https://hippogriff.github.io/CSGNet/}.

% -----------------------------------------------------------------------
\section{Related Work}

CSG parsing has a long history and a number of approaches have been
proposed in the literature over the past 20 years. 
Much of the earlier work can be categorized as ``bottom-up'' and
focuses on the problem of converting a boundary representation (b-Rep)
of the shape to a CSG program.
Our work is more related to program generation approaches using neural
networks which have recently seen a revival in the context of natural
language, graphics, and visual reasoning tasks. 
We briefly summarize prior work below.

%Our work is primarily related to neural program induction methods. Secondly, it
%is also related to ``vision-as-inverse-graphics'' approaches, as well as neural
%network-based methods that predict shape primitives or parameters of procedural
%graphics models. Below, we briefly overview these prior methods, and explain
%differences from our work.

\subsection{Bottom-up shape parsing} 
%Our work is related to bottom-up techniques for shape
%parsing using grammars.
%~\cite{fischler1973representation,felzenszwalb2005pictorial,yang2011articulated,bourdev2010detecting,Bokeloh:2010:CPS,TeboulKSKP11,Martinovic:2013:BGL,Talton:2012:LDP,ritchie2015controlling}.
An early example of a grammar-based shape parsing approach is the ``pictorial structure''
model~\cite{fischler1973representation}.
It uses a tree-structured
grammar to represent articulated objects and has been applied to
parsing and detecting humans and other
categories~\cite{felzenszwalb2005pictorial,yang2011articulated,bourdev2010detecting}.
However, the parse trees are often shallow and these methods rely on
accurate bottom-up proposals to guide parsing (\eg, face and
upper-body detection for humans). 
In contrast, primitive detection for CSG parsing is challenging as
shapes change significantly when boolean operations are applied to
them.
Approaches, such as~\cite{Shapiro:1991:COC,Buchele:2003:THC,Shapiro:1993:SBC}, assume an exact boundary
representation of primitives which is challenging to estimate from
noisy or low-resolution shapes.
This combined with the fact that parse trees for CSG can be
significantly deeper makes
bottom-up parsing error prone. Evolutionary approaches have also been investigated for optimizing CSG\ trees \cite{Hamza,Weiss,Fayolle}, however, they are computationally  expensive.

Thus, recent work has focused on reducing the complexity of
search.
Tao~\etal~\cite{Tao} directly operates on input meshes, and
converts the mixed domain of CSG trees (discrete
operations and continuous primitive locations) to a discrete domain that is
suitable for boolean satisfiability (SAT) based program synthesizers.
This is different from our approach which uses a neural network to
generate programs without relying on an external optimizer. 

%Wu~\etal~\cite{Wu} use a bottom-up procedure to parse the
%shapes represented in point cloud. 
%They extract a large number of primitive
%proposal and infer a CSG tree. While inferring the CSG tree they
%generate binary labels (inside/outside) for every bounding box for a primitive.
%A CSG tree is then constructed in a bottom up manner by minimizing the
%reconstruction error. \todo{What is so special about this?}
%\todo{I commented this one since it appears later.}

\subsection{Inverse procedural modeling } 
A popular approach to generate 3D shapes and scenes is to
infer context-free, often probabilistic ``shape grammars'' from a
small set of exemplars, then sample grammar derivations to create new
shapes\cite{vanegas2012inverse,stava2014inverse,ritchie2015controlling,Talton:2012:LDP}.
This approach called Inverse Procedural Modeling (IPM) has also been
used in analysis-by-synthesis image parsing frameworks
\cite{yuillek06,TeboulKSKP11,Martinovic:2013:BGL}.

%Shape grammars have the disadvantage of not modeling
%long-range dependencies in the parsing task, and are often specific to a
%particular shape class (\eg, buildings).\\
%A well-known approach to visual
%analysis is to generate and fit hypotheses of scenes or objects to input image
%data i.e., perform analysis-by-synthesis \cite{yuillek06}. 
%For example, Kulkani \etal \cite{Kulkarni2015inverse} proposed sampling-based probabilistic inference to
%estimate parameters of stochastic graphics models (e.g., human bodies,
%or rotationally symmetric objects) representing the space of
%hypothesized scenes given an input image. 
Recent approaches employ CNNs to infer parameters of objects
\cite{kulkarni2015deep} or whole scenes \cite{Romaszko2017inverse} to
aid procedural modeling. 
A similar trend is observed in graphics applications where CNNs are
used to map input images or partial shapes to procedural model parameters
\cite{huang2017shape,RitchieTHG16,Nishida:2016:ISU}. 
Wu \etal~\cite{JiajunWu} detect objects in scenes by employing a network for
producing object proposals and a network that predicts whether there
is an object in a proposed segment, along with various object attributes. 
Eslami \etal \cite{Eslami16} use a recurrent neural network to attend
to one object at a time in a scene, and learn to use an appropriate
number of inference steps to recover object counts, identities and poses. 

Our goal is fundamentally different: given a generic grammar
describing 2D or 3D modeling instructions and a target image or shape,
our method infers a derivation, or more specifically a modeling
program, that describes it. 
The underlying grammar for CSG is quite generic compared
to specialized shape grammars.
It can model shapes in several different classes
and domains (\eg, furniture, logos, \etc). 
%In contrast to IPM approaches, incorporating some notion of memory to
%capture long-range interactions between modeling commands is crucial
%element in our setting. 
%Futhermore, IPM methods attempt to fit grammar parameters to consistently parameterized input shape
%structures (e.g., shapes with labeled parts). Small parameter changes may yield
%drastically different visual results. 
%We also do not assume labeled shapes as
%input, and we use a rendering module to receive explicit visual feedback during
%learning to predict modeling instructions more accurately.

%In contrast, we do not aim at parsing images or scenes into a collection of
%objects and their parameters. We instead parse input imag or 3D shapes into a
%sequence of modeling operations on primitives (i.e, a visual program) to match a
%target image. In our setting, the space of outputs is much larger and the order
%of operations in our visual programs matter.

\subsection{Neural program induction} 
Our approach is inspired by recent work in using neural networks to
infer programs expressed in some high-level language, \eg, to answer question involving complex
arithmetic, logical, or semantic parsing operations \cite{Neelakantan,Reed2015,Denil,Balog,Joulin2015a,Zaremba2014,Zarembaa,Kaiser2015,Liang2016}.
Approaches, such as~\cite{Johnson,hu2017learning}, produce programs
composed of functions that perform compositional reasoning on an
image using an execution engine consisting of neural
modules~\cite{Andreas2016NeuralMN}.
Similarly, our method produces a program consisting
of shape modeling instructions to match a target image by
incorporating a shape renderer.
%execution, a procedure to train our architecture on automatically generated
%synthetic programs, and a RL-based procedure that uses our renderer to provide
%visual feedback during training.

Other related work include the recent work by Tian~\etal
\cite{tian2018learning}, which proposes a program induction
architecture for 3D shape modeling. Here programs contain a variety of 
primitives and symmetries are incorporated with loops. 
%The
%training is done through a combination of supervised learning with ground
%truth programs and self-supervised learning with a differentiable
%program execution module. 
While this is effective for categories such as
chairs, the lack of boolean operations is limiting.
A more complex approach is that of Ellis \etal \cite{Ellis},
who synthesize hand-drawn shapes by combining (lines, circles,
rectangles) into Latex programs.
Program synthesis is posed as a constraint satisfaction problem which is
computationally expensive and can take hours to solve. 
In contrast, our feed-forward model that takes a
fraction of a second to generate a program.

\subsection{Primitive fitting}

Deep networks have recently been applied to a wide range of primitive
fitting problems for 2D and 3D shapes. 
Tulsiani \etal
\cite{abstractionTulsiani17} proposed a volumetric CNN
that predicts a fixed number of cuboidal primitives to describe an
input 3D shape. 
Zou \etal \cite{Zou} proposed an LSTM-based architecture to predict a
variable number of boxes given input depth images. 
%Our approach on the other hand models richer primitives and their
%boolean combinations, supporting a richer modeling paradigm.
%limited to a single type of primitives (e.g., cubes), and additionalyl
%outputs modeling operations acting on them, or in other words supports a
%significantly richer modeling paradigm. The program can be used not only to
%geometrically describe the input shape but can also be directly edited to
%manipulate it if desired. 
Li \etal \cite{Lingxiao} introduced a point cloud based primitive
fitting network where shapes are represented as an union of
primitives.
Paschalidou \etal \cite{Paschalidou} uses superquadrics instead of
traditional cuboids.
Huang~\etal~\cite{Huang2018} decompose an image by
detecting primitives and arranging them into layers.
Gao~\etal \cite{Gao2019} train deep network to produce control points
for splines using input images and point cloud.
The above approaches
are trained to minimize
reconstruction error like our method. On the other hand, they are limited to primitive fitting, while our method also learns modeling operating on them.

% to 
%primitives, which with the help of chamfer distance based loss function learn to
%represent 3D shapes without explicit supervision. 

%\subsection{Memory Networks}
%Memory based networks have applications in question answering, machine
%translation, and tasks that require long range reasoning. Joulin \etal
%\cite{Joulin2015a} introduced Stack-RNN, that simulates an algorithmic stack in
%the form of structured memory in the network. The external memory stores an
%embedding representing occupancy of elements in the stack, which gives the
%network the capacity to count and memorize the sequence. On the contrary, we
%make use of the CSG renderer to provide an explicit execution stack as input to
%the network, where each element in the stack is an intermediate shape produced
%by the renderer while executing the program. In this way, structure in the
%network is also imposed by the structure of the stack. Using an explicit stack
%as input removes the need to learn an abstract representation of the stack
%within the network. Several authors \cite{Ganin,Zou,Ellis} also use the
%intermediate output of the execution engine in the neural network to better
%react to the consequence of its produced actions.

% -----------------------------------------------------------------------
\section{Designing a Neural Shape Parser} \label{shape-parsing}
 \begin{figure*}
\centering
\includegraphics[width=0.7\linewidth]{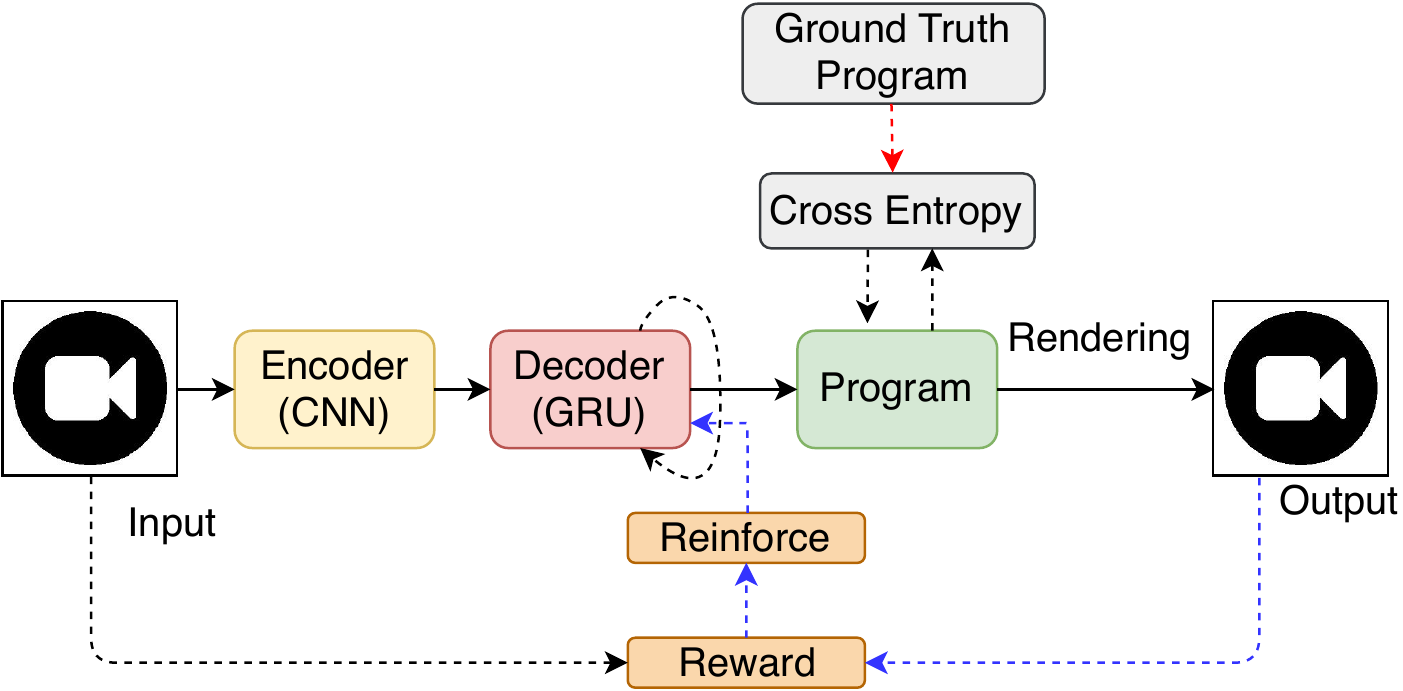}
\vskip -2mm    
\caption{\label{fig:overview}\textbf{Overview of our approach.} Our
  neural shape parser consists of two parts: first at every time step encoder
  takes as input a target shape (2D or 3D) and outputs a feature vector through
  CNN. Second, a decoder maps these features to a sequence of modeling
  instructions yielding a visual program. The rendering engine processes
  the program and outputs the final shape. The training signal can either come from
  ground truth programs when such are available, or in the form of rewards after
  rendering the predicted programs.}
\end{figure*}

 \begin{figure*}[!htbp]
\centering
\includegraphics[width=\textwidth]{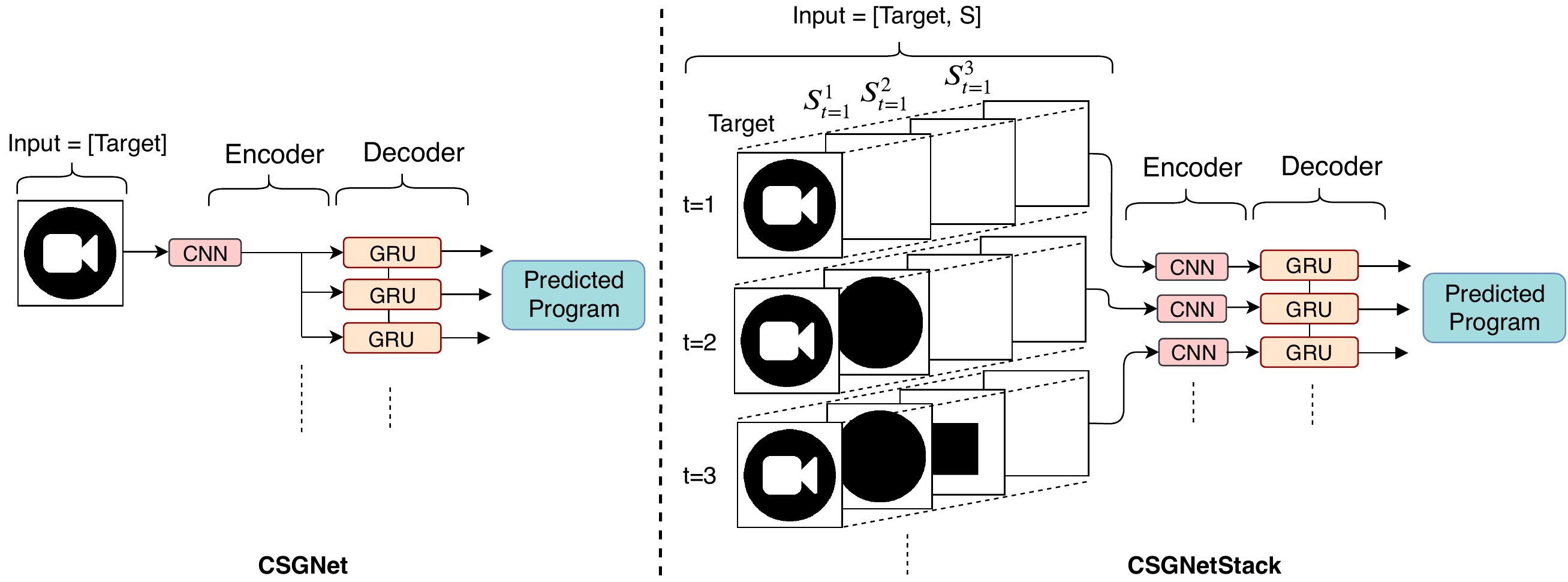}
\vskip -2mm    
\caption{\label{fig:arch}\textbf{Two proposed architectures of our
    neural shape parser \csgnet (left), \scsgnet (right)}. CSGNet takes
the  target shape as input and encodes it using a CNN, whereas in \scsgnet, the target shape is concatenated with stack $S_t$ along the
  channel dimension and passes as input to the CNN encoder at
  every time step. Empty entries in the stack
  are shown in white.}
\vskip -3mm
\end{figure*}

In this section, we first present a neural shape parser, called \csgnet, that  induces
programs based on a CSG grammar given only 2D/3D shapes as input. We also present another shape parser variant, called \scsgnet,  which incorporates a stack as a form of explicit memory and results in improved accuracy and faster training. We show that both variants can be
trained to produce CSG programs in a supervised learning setting when ground-truth
programs are available. When these are not available, we show that reinforcement learning can be used based on  policy gradient and reward shaping
techniques. Finally, we describe ways to improve the shape parsing at test time through a post-processing stage.\\

\noindent
\textbf{\csgnet}. The goal of a \textbf{shape parser} $\pi$ is to produce a
sequence of instructions given an input shape. The parser can be implemented as
an encoder-decoder using neural network modules as shown in
Figure~\ref{fig:overview}. The \textbf{encoder} takes as input an image $I$ and
produces an encoding $\Phi(I)$ using a CNN. The \textbf{decoder} $\Theta$ takes
as input $\Phi(I)$ and produces a probability distribution over programs $P$
represented as a sequence of instructions. Decoders can be implemented using
Recurrent Neural Networks (RNNs). We employ Gated Recurrent Units
(GRUs)~\cite{chung2014empirical} that have been widely used for sequence
prediction tasks such as generating natural language and speech. The overall
network can be written as $\pi(I) = \Theta \circ \Phi (I)$. We call this basic
architecture as \csgnet (see also Figure \ref{fig:arch}, left).\\

\noindent
\textbf{\scsgnet.} The above architecture can further be improved by
incorporating feedback from the renderer back to the network. More specifically,
the encoder can be augmented with an execution stack that stores the result of
the renderer at every time step along with the input shape. This enables the
network to adapt to the current rendered result. To accomplish this, our CSG
rendering engine executes the program instructions produced by the decoder with
the help of stack $S=\{s_t:t=1,2 \ldots \}$ at each time step $t$. The stack is
updated after every instruction is executed and contains intermediate shapes
produced by previous boolean operations or simply an initially drawn shape
primitive. This stack of shapes is concatenated with the target shape, all
stored as binary maps, along the channel dimension. The concatenated map is
processed by the network at the next time step. Instead of taking all elements
of the stack, which vary in number depending on the generated program, we only
take the top-$K$ maps of the stack. Empty entries in the stack are represented
as all-zero maps (see also Figure \ref{fig:arch}, right). At the first time
step, the stack is empty, so all $K$ maps are zero.

In our implementation, the parser $\pi$ takes $Z=[I, S]$ as input of size
$64\times 64 \times (K+1)$ for $2D$ networks and $64 \times 64 \times 64 \times
(K+1)$ for $3D$ networks, where $I$ is the input shape, $S$ is the execution
stack of the renderer, and $K$ is the size of the stack. The number of channels
is $(K+1)$\ since the target shape, also represented as $64^2$ (or $64^3$ in
3D), is concatenated with the stack. Details of the architecture are described
in Section \ref{experiment}. Similarly to the basic \csgnet architecture, the
encoder takes $Z$ as input and yields a fixed length encoding $\Phi(Z)$, which
is passed as input to the decoder $\Theta$ to produce a probability distribution
over programs $P$. The stack-based network can be written as $\pi(Z) = \Theta
\circ \Phi (Z)$. We call this stack based architecture \scsgnet. The difference
between the two architectures is illustrated in Figure \ref{fig:arch}.

\noindent
\textbf{Grammar.} The space of programs can be efficiently described according to
a context-free grammar~\cite{hopcroft}. For example, in constructive solid
geometry the instructions consist of drawing primitives (eg, spheres, cubes,
cylinders, etc) and performing boolean operations described as a grammar with
the following production rules:

\begin{align*}
&S \rightarrow E \\
&E \rightarrow E~E~T ~|~ P \\
&T \rightarrow \texttt{OP}_1 |\texttt{OP}_2 | \ldots |  \texttt{OP}_m  \\
&P \rightarrow \texttt{SHAPE}_1|\texttt{SHAPE}_2| \ldots | \texttt{SHAPE}_n
\end{align*}

\begin{figure*}
  \renewcommand{\arraystretch}{1.2}
  \begin{tabular}{l|l|l|c}
    \textbf{Instruction} & \textbf{Execution} & \textbf{Stack} & \multirow{7}{*}{\includegraphics[scale=.42]{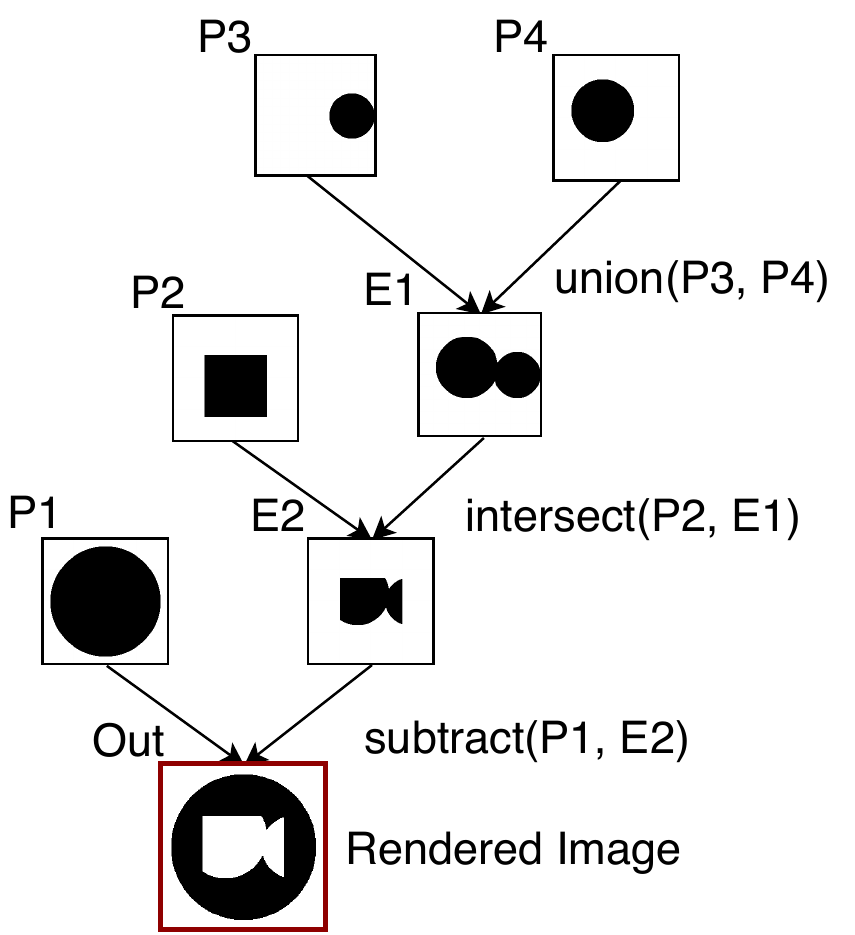}} \\
    \cline{1-3}
    \texttt{circle(32,32,28)} & \texttt{push circle(32,32,28)} & \texttt{[P1]} & \\
    \texttt{square(32,40,24)} & \texttt{push square(32,40,24)} & \texttt{[P2 P1]} & \\
    \texttt{circle(48,32,12)} & \texttt{push circle(48,32,12)} & \texttt{[P3 P2 P1]} & \\
    \texttt{circle(24,32,16)} & \texttt{push circle(24,32,16)} & \texttt{[P4 P3 P2 P1]} & \\
    \texttt{union} & \texttt{A=pop; B=pop; push(B$\cup$A)}     & \texttt{[E1 P2 P1] // E1=P3$\cup$P4} & \\
    \texttt{intersect} & \texttt{A=pop; B=pop; push(B$\cap$A)} & \texttt{[E2 P1] // E2=P2$\cap$E1} & \\
    \texttt{subtract} & \texttt{A=pop; B=pop; push(B-A)} & \texttt{[Out] // Out=P1-E2} & \\
  \end{tabular}
  \caption{\textbf{Example program execution.} Each row in the table from the top shows the instructions, program execution, and the current state of the stack of the shift-reduce CSG parser. On the right is a graphical representation of the program. An instruction corresponding to a primitive leads to \texttt{push} operation on the stack, while an operator instruction results in popping the top two elements of the stack and pushing the result of applying this operator.}
  \label{table:instructions}
\end{figure*}

Each rule indicates possible derivations of a non-terminal symbol separated by
the $|$ symbol. Here $S$ is the start symbol, $\texttt{OP}_i$ is chosen from a
set of defined modeling operations and the $\texttt{SHAPE}_i$ is a primitive
chosen from a set of basic shapes at different positions, scales, orientations,
etc. Instructions can be written in a standard post-fix notation, \eg, 
$\texttt{SHAPE}_1\texttt{SHAPE}_2\texttt{OP}_1\texttt{SHAPE}_3\texttt{OP}_2$.
Table \ref{table:instructions} shows an example of a program predicted by the
network and corresponding rendering process.

\subsection{Learning}\label{learning}
Given the input shape $I$ and execution stack $S$ of the renderer, the parser
network $\pi$ generates a program that minimizes a reconstruction error between
the shape produced by executing the program and a target shape. Note that not
all programs are valid hence the network must also learn to generate grammatical
programs.
\subsubsection{Supervised learning} \label{supervised-learning} When target
programs are available both \csgnet and \scsgnet variants can be trained with
standard supervised learning techniques. Training data consists of $N$ shapes,
$P$ corresponding programs, and also in the case of \scsgnet $S$ stacks, 
program triplets $(I^i, S^i, P^i)$, $i=1, \ldots, N$. The ground-truth program
$P^i$ can be written as a sequence of instructions $g^i_1$, $g^i_2$ ..
$g^i_{T_i}$, where $T_i$ is the length of the program $P^i$. Similarly, in the
case of \scsgnet, the $S^i$ can be written as sequence of states of stack
$s^i_1$, $s^i_2$ .. $s^i_{T_i}$ used by the rendering engine while executing the
instructions in program $P^i$. Note that while training in supervised setting,
the stack $s_t$ is generated by the renderer while executing ground truth
instructions $g_{1:t}$, but during inference time, the stack is generated by the
renderer while executing the predicted instructions. For both network variants,
the RNN produces a categorical distribution $\pi$ for both variants.

The parameters $\theta$ for either variant can be learned to maximize the
log-likelihood of the ground truth instructions:
\begin{equation}
{\cal L}(\theta) = \sum_{i=1}^{N} \sum_{t=1}^{T_i} \log \pi_\theta(g^i_t|g^i_{1:t-1},s^i_{1:t-1},I^i)
\end{equation}

\subsubsection{Learning with policy gradients} \label{policy-gradient} 
Without target programs one can minimize a reconstruction error between the shape
obtained by executing the program and the target. However, directly minimizing
this error using gradient-based techniques is not possible since the output
space is discrete and execution engines are typically not differentiable. Policy
gradient techniques~\cite{Williams92simplestatistical} from the reinforcement
learning (RL) literature can instead be used in this case.

Concretely, the parser $\pi_{\theta}$, that represents a policy network, can be
used to sample a program $y$ = ($a_1$,$a_2$ .. $a_T$) conditioned on the input
shape $I,$ and in the case of \scsgnet, also on the stack $S$ = ($s_1$, $s_2$ ..
$s_{T}$). Note that while training using policy gradient and during inference
time, the stack $s_t$ is generated by the renderer while executing predicted
instructions by the parser since ground-truth programs are unavailable. Then a
reward $R$ can be estimated by measuring the similarity between the generated
image $\hat{I}$ obtained by executing the program and the target shape $I$. With
this setup, we want to learn the network parameters $\theta$ that maximize the
expected rewards over programs sampled under the predicted distribution
$\pi_{\theta}(y|S, I)$ across images $I$ sampled from a distribution ${\cal D}$:
\begin{equation*} \label{expected-return}
\mathbb{E}_{I \sim {\cal D}} \left[ J_\theta(I) \right] = \mathbb{E}_{I \sim {\cal D}} \sum^T_{t=1}\mathbb{E}_{y_t\sim \pi_\theta(y|s_{1:t-1}, I)}\left[ R\right]
\end{equation*}

The outer expectation can be replaced by a sample estimate on the training data.
The gradient of the inner expectation can be obtained by rearranging the
equation as\footnote{conditioning on stack and input image is removed for the
  sake of brevity.}:

\begin{equation*} \label{expected-grad}
\nabla_{\theta} J_\theta(I)= \nabla_{\theta} \sum_y \pi_\theta(y) R  = \sum_y \nabla_{\theta} \log \pi_\theta(y) \left[\pi_\theta(y) R \right]
\end{equation*}

It is often intractable to compute the expectation $J_\theta(I)$ since the space
of programs is very large. Hence the expectation must be approximated. The
popular REINFORCE~\cite{Williams92simplestatistical} algorithm computes a
Monte-Carlo estimate as:
\begin{equation*}
\nabla_{\theta} J_\theta(I) = \frac{1}{M} \sum_{m=1}^{M} \sum_{t=1}^{T} \nabla \log \pi_\theta (\hat{a}_t^m| \hat{a}_{1:t-1}^m, \hat{s}^m_{1:t-1}, I)R^m 
\end{equation*}
by sampling $M$ programs from the policy $\pi_{\theta}$. Each program $y^m$ is
obtained by sampling instructions $\hat{a}^m_{t=1:T}$ from the distribution
$\hat{a}_t^m\sim\pi_{\theta}(a_t|\hat{a}^m_{1:t-1};\hat{s}^m_{1:t-1}, I)$ at every time step $t$ until the stop symbol (\texttt{EOS}) is sampled. The reward $R^m$ is calculated
by executing the program $y^m$. Sampling-based estimates typically have high
variance that can be reduced by subtracting a baseline without changing the bias
as:
\begin{equation}
\nabla_{\theta} J_\theta(I) \!=\! \frac{1}{M} \sum_{m=1}^{M} \sum_{t=1}^{T} \nabla_{\theta} \log \pi_{\theta} (\hat{a}_{t}^m| \hat{a}_{1:t-1}^m, \hat{s}^m_{1:t-1}, I)(R^m - b)
\end{equation}
A good choice of the baseline is the expected value of returns starting from $t$
\cite{Sutton:1999:PGM:3009657.3009806,Williams92simplestatistical}. We compute the
baseline as the running average of past rewards.\\

\noindent
\textbf{Reward.} The rewards should be primarily designed to encourage visual
similarity of the generated program with the target. Visual similarity between
two shapes is measured using the Chamfer distance (CD) between points on the
edges of each shape. The CD is between two point sets, $\mathbf{x}$ and
$\mathbf{y}$, is defined as follows:
\begin{equation}
  Ch(\mathbf{x}, \mathbf{y})=
           \frac{1}{2|\mathbf{x}|}\sum_{x \in \mathbf{x}} \min_{y \in \mathbf{y}} \norm{x-y}_2 +
           \frac{1}{2|\mathbf{y}|}\sum_{y \in \mathbf{y}} \min_{x \in \mathbf{x}} \norm{x-y}_2 \nonumber
\end{equation}
The points are scaled by the image diagonal, thus $Ch(\mathbf{x}, \mathbf{y})
\in [0,1]~\forall \mathbf{x}, \mathbf{y}$. The distance can be efficiently
computed using distance transforms. In our implementation, we also set a maximum
length $T$ for the induced programs to avoid having too long or redundant
programs (\eg, repeating the same modeling instructions over and over again).
We then define the reward as:
\begin{equation*}
R =  \begin{cases}
f\left(Ch(\texttt{Edge}(I),\texttt{Edge}(\Re(y)\right), &y \text{ is valid} \\
0,          &y \text{ is invalid} \\
\end{cases}
\end{equation*}
where $f$ is a reward shaping function and $\Re$ is the CSG rendering engine
that renders the program $y$ into a binary image. Since invalid programs get
zero reward, the maximum length constraint on the programs encourages the
network to produce shorter programs with high rewards. We use maximum length
$T=13$ in all of our RL experiments. The function $f$ shapes the CD as $f(x) =
(1-x)^\gamma$ with an exponent $\gamma > 0$. Higher values of $\gamma$ makes the
reward closer to zero, thereby making the network to produce programs with
smaller CD. Table \ref{table:gamma} (left) shows the dynamics of reward shaping
function with different $\gamma$ value and (right) shows that increasing
$\gamma$ values decreases the average CD calculated over the test set. We choose
$\gamma=20$ in our experiments.

\begin{table}[]
\begin{tabular}{ll}
\includegraphics[width=0.48\linewidth]{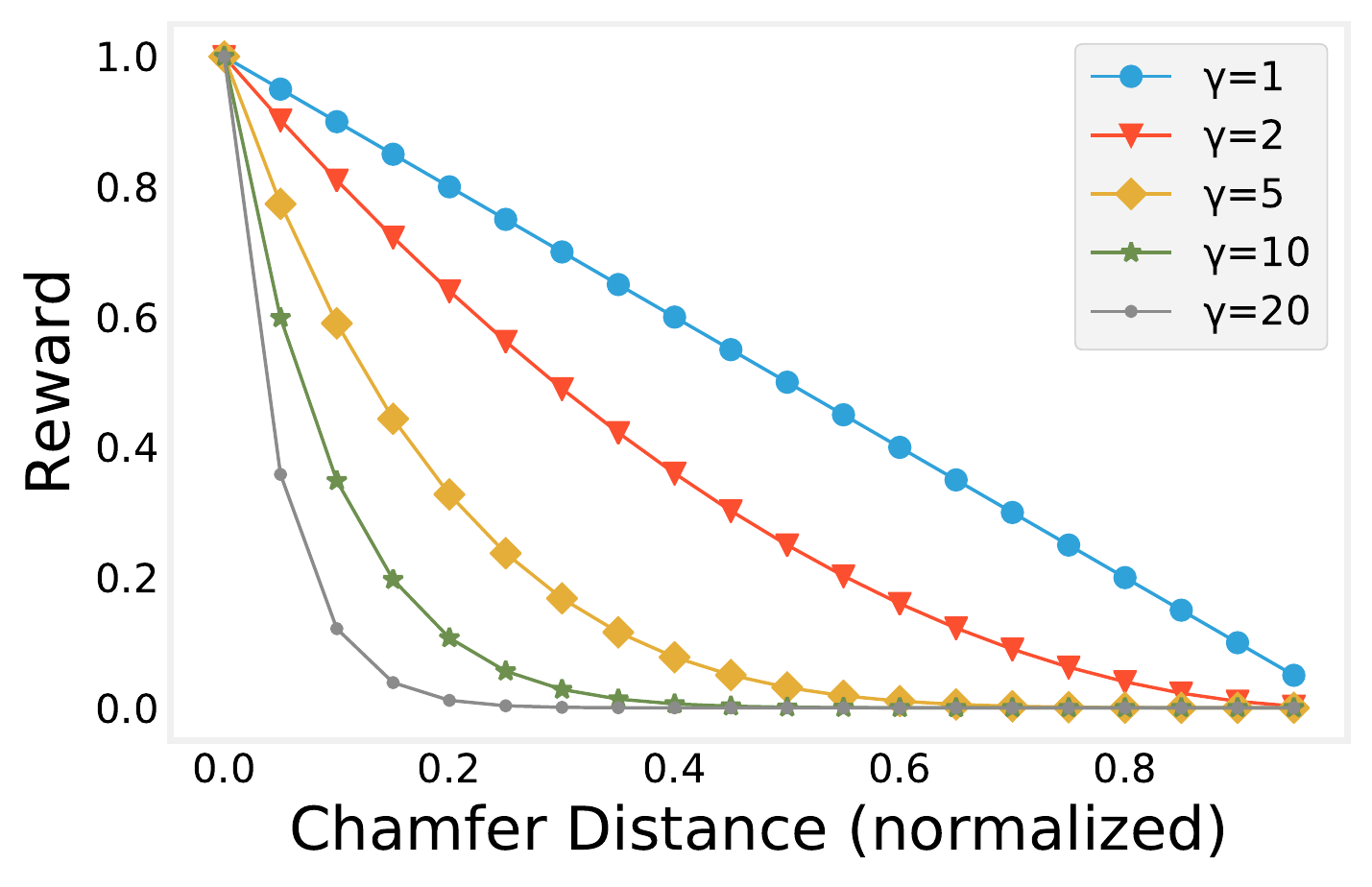}&
\includegraphics[width=0.48\linewidth]{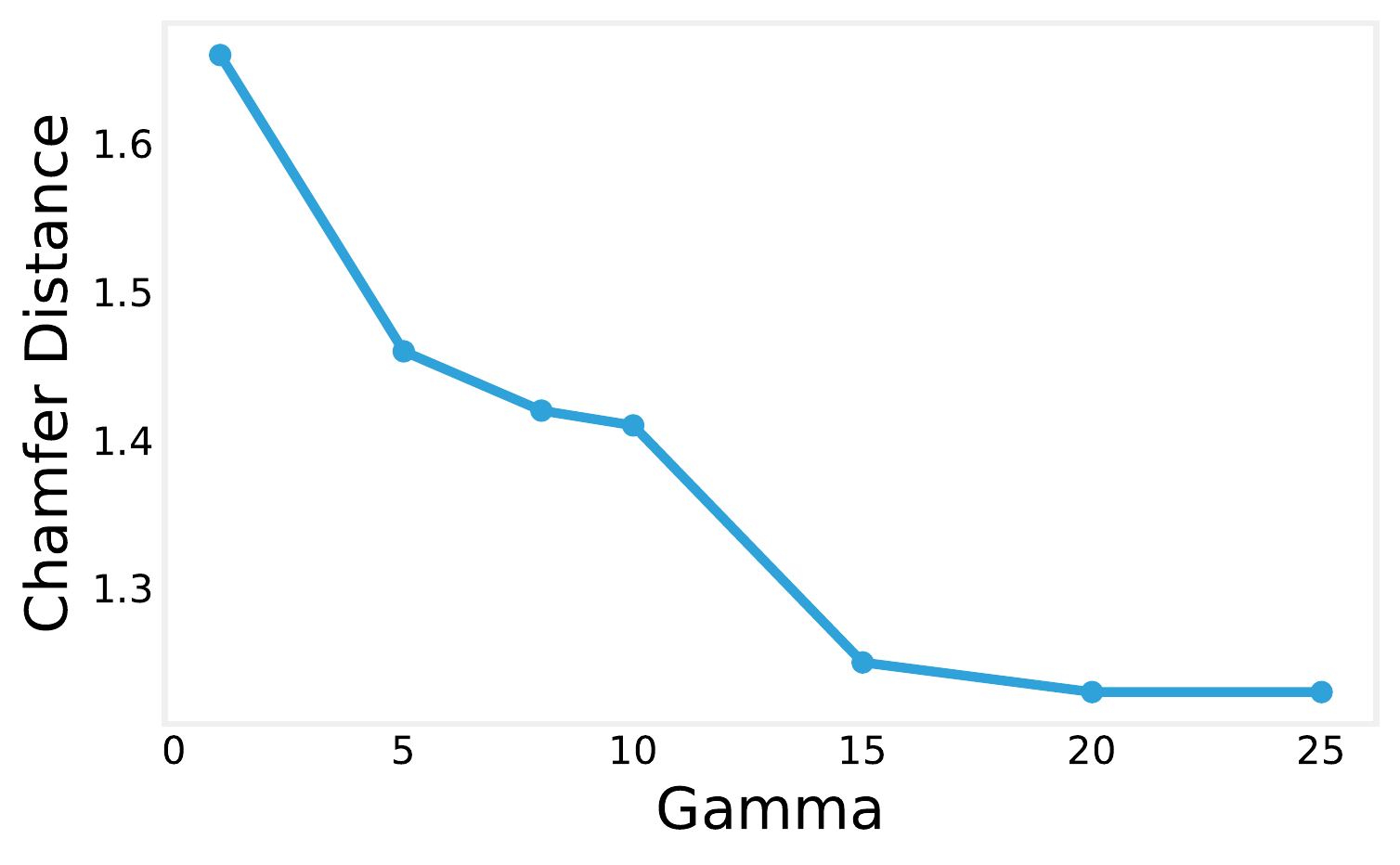}
\end{tabular}
\caption{\textbf{Reward shaping.} (Left) We visualize the skewness introduced by
  the $\gamma$ in the reward function. (Right) Larger $\gamma$ value produces
  smaller CD (in number of pixels) when our model is trained using REINFORCE.}
\label{table:gamma}
\end{table}

\subsection{Inference} \label{inference}
\textbf{Greedy decoding and beam search.} Estimating the most likely
program given an input is intractable using RNNs. Instead one usually employs a
greedy decoder that picks the most likely instruction at each time step. An
alternate is to use a beam search procedure that maintains the k-best likely
sequences at each time step. In our experiments we report results with varying
beam sizes.\\

\noindent
\textbf{Visually-guided refinement.} Both parser variants produce a program with
a discrete set of primitives. However, further refinement can be done by
directly optimizing the position and size of the primitives to maximize the
reward. The refinement step keeps the program structure of the program and
primitive type fixed but uses a heuristic algorithm~\cite{Powell1964} to
optimize the parameters using feedback from the rendering engine. In our
experiments, we observed that the algorithm converges to a local minima in about
$10$ iterations and consistently improves the results.

% -----------------------------------------------------------------------
\section{Experiments} \label{experiment}
\begin{figure}[]
        \centering
        \includegraphics[width=\linewidth]{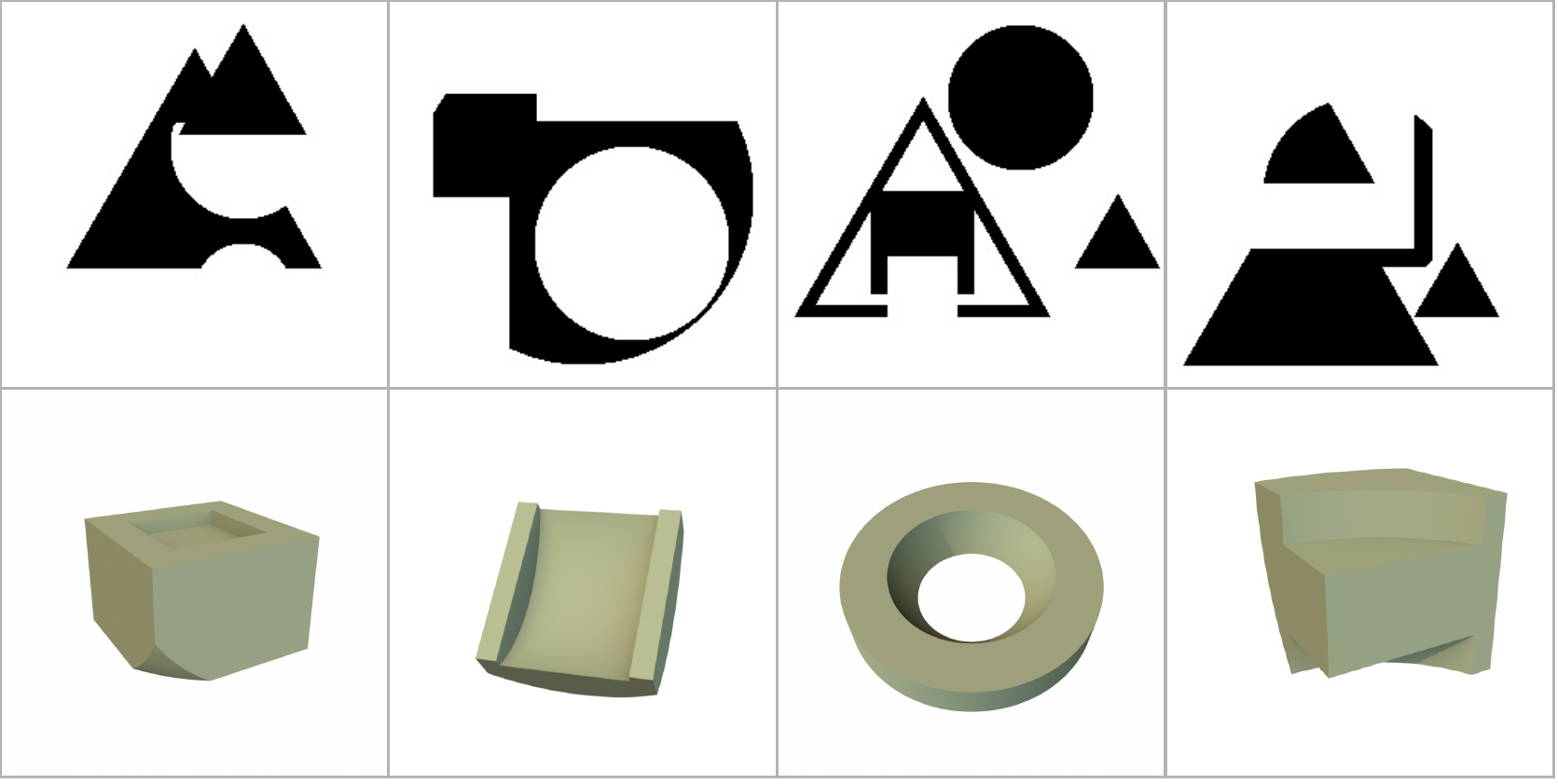}
        \caption{\textbf{Samples of our synthetically generated
            programs.} 2D\ samples are in the top row and 3D\ samples in the bottom.}
        \label{fig:samples}
\end{figure}
We describe our experiments on different datasets exploring the generalization
capabilities of our network variants (\csgnet and \scsgnet). We first describe
our datasets: (i)\ an automatically generated dataset of 2D and 3D\ shapes based
on synthetic generation of CSG\ programs, (ii) 2D\ CAD\ shapes mined from the
web where ground-truth programs are not available, and (iii) logo images mined
also from the web where ground-truth programs are also not available. Below we discuss
our qualitative and quantitative results on the above dataset.

\subsection{Datasets} \label{grammar}
To train our network in the supervised learning setting, we automatically
created a large set of 2D\ and 3D CSG-based synthetic programs according to the
grammars described below.\\

\noindent
\textbf{Synthetic 2D shapes.} We sampled derivations of the
following CSG grammar to create our synthetic dataset in the 2D\ case:
\begin{align*}
&S \rightarrow E;\\
&E \rightarrow EET \mid P(L, R);\\
&T \rightarrow intersect \mid union \mid subtract;  \\
&P \rightarrow square \mid circle  \mid triangle;  \\
&L \rightarrow \big[8:8:56\big]^2;~~ R \rightarrow \big[8:4:32\big].
\end{align*}
Primitives are specified by their type: \emph{square}, \emph{circle}, or
\emph{triangle}, locations $L$ and circumscribing circle of radius $R$ on a
canvas of size $64\times64$. There are three boolean operations: $intersect$,
$union$, and $subtract$. L is discretized to lie on a square grid with spacing
of $8$ units and R is discretized with spacing of $4$ units. The
\emph{triangles} are assumed to be upright and equilateral. The synthetic
dataset is created by sampling random programs containing different number of
primitives from the above grammar, constraining the distribution of various
primitive types and operation types to be uniform. We also ensure that no
duplicate programs exist in our dataset. The primitives are rendered as binary
images and the programs are executed on a canvas of $64\times64$ pixels. Samples from
our dataset are shown in Figure \ref{fig:samples}. Table \ref{table:dataset}
provides details about the size and splits of our dataset. \\

\begin{table}
  \centering
  \setlength{\tabcolsep}{3pt}
\begin{tabular}{c|c|c|c|c|c|c}
\multirow{2}{*}{\begin{tabular}[c]{@{}c@{}}Program\\ Length\end{tabular}} & \multicolumn{3}{c|}{2D} & \multicolumn{3}{c}{3D} \\
\cline{2-7}
                              & Train   & Val  & Test  & Train   & Val  & Test  \\
\hline
3                             & 25k     & 5k   & 5k    & 100k    & 10k  & 20k   \\
5                             & 100k    & 10k  & 50k   & 200k    & 20k  & 40k   \\
7                             & 150k    & 20k  & 50k   & 400k    & 40k  & 80k   \\
9                             & 250k    & 20k  & 50k   & -       & -    & -     \\
11                            & 350k    & 20k  & 100k  & -       & -    & -     \\
13                            & 350k    & 20k  & 100k  & -       & -    & -    
\end{tabular}
\caption{\textbf{Statistics of our 2D and 3D synthetic dataset.}}
  \label{table:dataset}
\end{table}
% While creating images, we impose following restrictions on shape primitives
% and operations: a) Primitives must lie completely inside the canvas, b) Each
% operation changes the number of ON pixels by atleast a set threshold. This
% avoids spurious operations, such as subtraction between shapes with little
% overlap. c) The number of ON pixels in the final image is above a threshold.
% The above scheme promotes programs with the Union operation. Hence, to ensure
% a balanced dataset, we boost the probabilities of generating programs with
% Difference and Intersection operations.

\noindent
\textbf{Synthetic 3D shapes.}
We sampled derivations of the following  grammar  in the case of 3D\ CSG:
\begin{align*}
  &S \rightarrow E; ~E \rightarrow EET; \\
  &E \rightarrow sp(L,R) \mid cu(L,R) \mid cy(L,R,H)\\
  &T \rightarrow intersect \mid union \mid subtract;  \\
  &L \rightarrow \big[8:8:56]^3\\
  &R \rightarrow \big[8:4:32];~H \rightarrow \big[8:4:32].
\end{align*}

The operations are same as in the 2D\ case. 
Three basic solids are denoted by `$sp$': Sphere, `$cu$': Cube, `$cy$': Cylinder. $L$ represents
the center of primitive in a 3D voxel grid. 
$R$ specifies radius of sphere and cylinder, or the size of cube. 
$H$ is the height of cylinder. The
primitives are rendered as voxels and the programs are executed on a 3D
volumetric grid of size $64$ $\times$ $64$ $\times$ $64$. We used the same
random sampling method as used for the synthetic 2D dataset, resulting in
3D\ CSG\ programs. 3D\ shape samples are shown in Figure
\ref{fig:samples}.\\

\noindent
\textbf{2D CAD shapes.}
We collected $8K$ CAD shapes from the Trimble 3DWarehouse
dataset~\cite{TrimbleWarehouse} in three categories: chair, desk and lamps. We
rendered the CAD\ shapes into $64$ $\times$ $64$ binary masks from their front and
side views. In Section \ref{experiment}, we show that the rendered shapes can be
parsed effectively through our visual program induction method. We split this
dataset into $5K$ shapes for training, $1.5K$ validation and $1.5K$ for testing.\\

\noindent
\textbf{Web logos.}
We mined $20$ binary logos  from the web that can be modeled using the primitives in our output shapes. We test our approach on these logos without further training or fine-tuning  our net on this data.
\begin{figure*}
\centering
\begin{subfigure}{.49\textwidth}
  \centering
  \includegraphics[width=\linewidth]{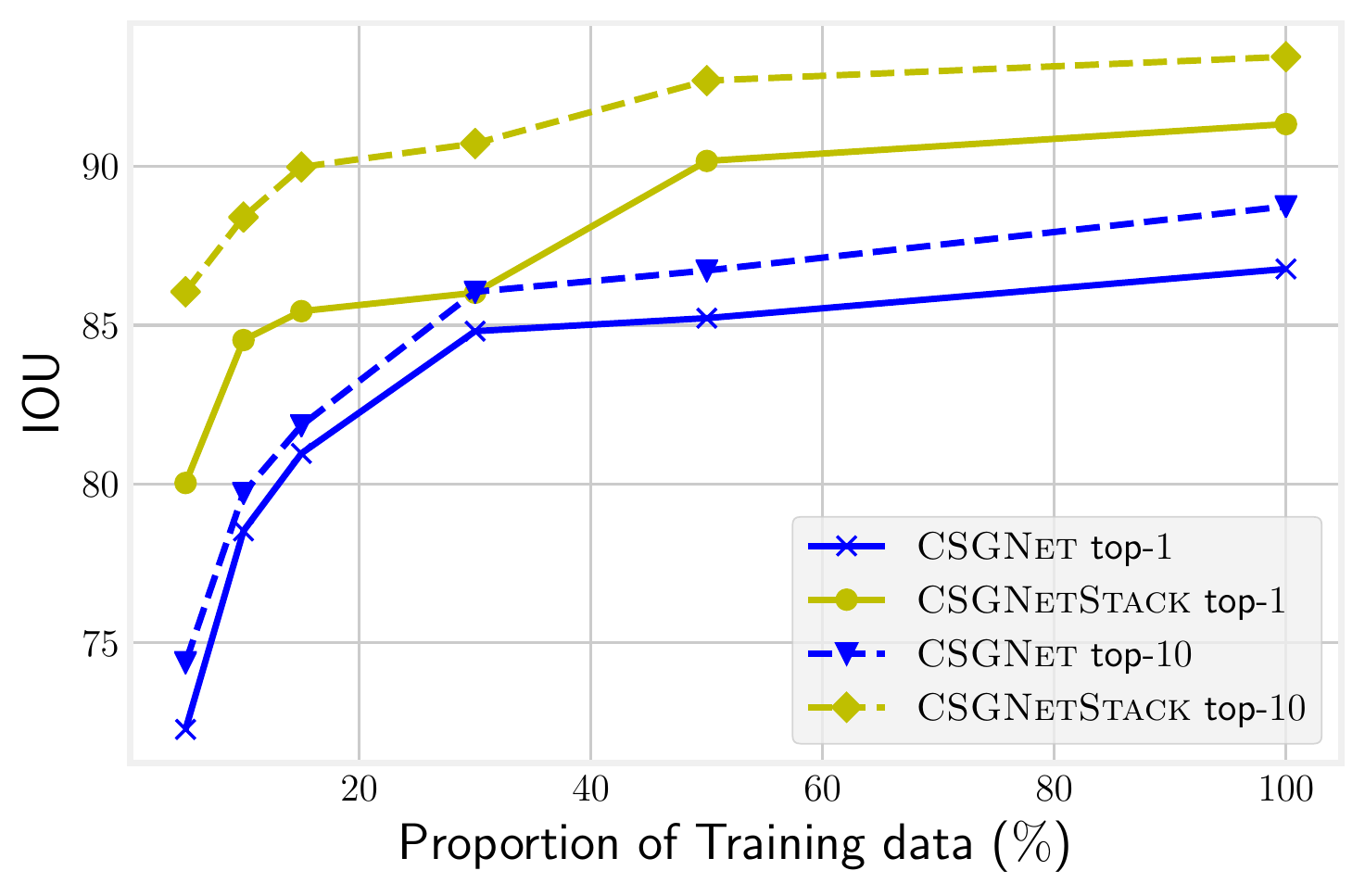}
\end{subfigure}
\begin{subfigure}{.49\textwidth}
  \centering
  \includegraphics[width=\linewidth]{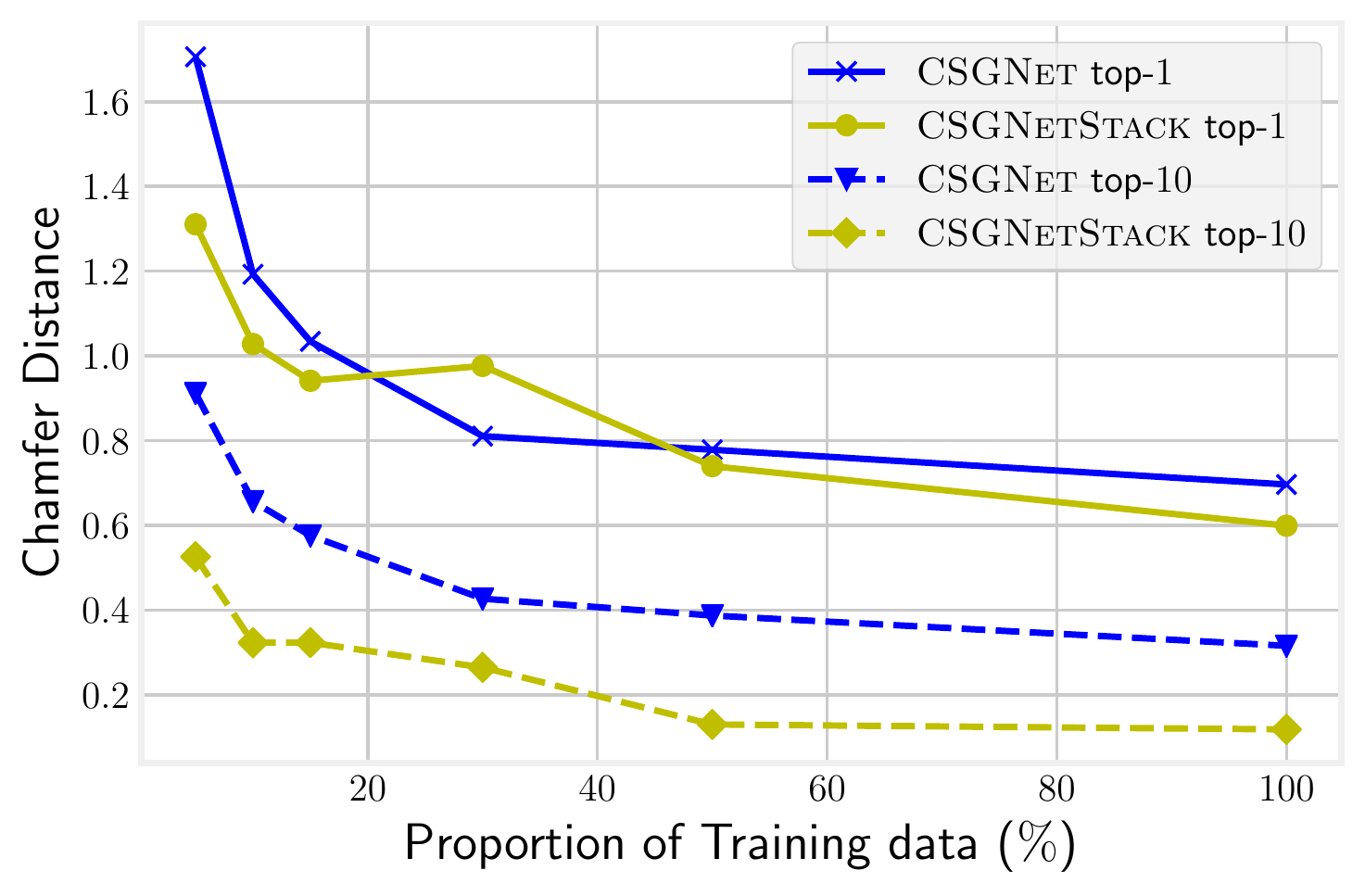}
\end{subfigure}
\caption{ \textbf{Performance (Left: IOU, Right: chamfer distance) of models
      by changing training size.} Training is done using $x\%$ of the complete
    dataset, where $x$ is shown on the horizontal axis. The top-$k$ beam sizes used during decoding at test time are shown in the legend. The performance of \csgnet (our basic non-stack neural shape parser) is shown in blue and the performance of \scsgnet (our variant that uses the execution stack) is shown in lime.}
\label{fig:synthetic-size}
\end{figure*}

\subsection{Implementation details}
\textbf{2D shape parsing.}
Our encoder is based on an image-based convnet in the case of 2D\
inputs. In the case of \scsgnet, the input to the network is a fixed size
stack along with target image concatenated along the channel dimension,  resulting in an
the input tensor of size $64 \times 64 \times (K + 1)$, where $K$ is the number of used maps in the stack (stack size). In the architecture without stack (\csgnet), $K$ is simply set to 0.
The output of the encoder is passed as input to our GRU-based decoder
at every program step. The hidden state of our GRU units is passed through two
fully-connected layers, which are then converted into a probability distribution
over program instructions through a classification layer. For the 2D CSG there
are $400$ unique instructions corresponding to $396$ different primitive types,
discrete locations and sizes, the $3$ boolean operations and the stop symbol.\\

\noindent
\textbf{3D\ shape parsing.} In the case of 3D shapes, the encoder is based on an
volumetric, voxel-based convnet. 3D-\scsgnet concatenates the stack with the
target shape along the channel dimension, resulting in an input tensor of size
$64\times 64 \times 64 \times (K + 1)$, where $K$ is the number of used maps in
the stack (stack size). In the architecture without stack (3D-\csgnet), $K$ is
simply set to 0. The encoder comprises of multiple layers of $3D$ convolutions
yielding a fixed size encoding vector. Similarly to the $2D$ case, the GRU-based
decoder takes the output of the encoder and sequentially produces the program
instructions. In this case, there are $6635$ unique instructions with $6631$
different types of primitives with different sizes and locations, plus $3$
boolean modeling operations and a stop symbol.

During training, on synthetic dataset, we sample images/3D shapes rendered from
programs of variable length (up to $13$ for 2D and up to $7$ for 3D dataset)
from training dataset from Table \ref{table:dataset}. More details about the
architecture of our encoder and decoder (number and type of layers) are provided
in the supplementary material.

% (2D CNN:
% \texttt{c:3x3x1x8  $\rightarrow$ c:3x3x8x16 $\rightarrow$ c:3x3x16x32}, 3D CNN: \texttt{c:4x4x4x1x32 $\rightarrow$
%   c:4x4x4x32x32 $\rightarrow$ c:4x4x4x32x64 $\rightarrow$ c:4x4x4x64x128 $\rightarrow$ c:4x4x4x128x256})

%architecture as used in supervised settings. We experiment with Intersection
%over Union (IOU) and Chamfer distance (CD) as reward function. We find that
%chamfer distance produces better qualitative results. Once the <EOS> is sampled
%from the policy distribution or maximum length ($T$)is achieved, validity of the
%program is checked based on the grammar defined in section \ref{language}. Valid
%programs are rendered using renderer $Z$ and reward is calculated based on some
%similarity metric $S$ (like IOU or CD). $S$ takes predicted shape, and target
%shape and gives a similarity score (reward r). Invalid program recieves zero
%reward. This is done to discourage network to generate invalid programs.
%  Note that valid program can still get zero reward if, for example, intersection
%  or subtraction is done between two disjoint shapes.

For supervised learning, we use the Adam optimizer~\cite{KingmaB14} with
learning rate $0.001$ and dropout of $0.2$ in non-recurrent network connections.
For reinforcement learning, we use stochastic gradient descent with $0.9$
momentum, $0.01$ learning rate, and with the same dropout as above.

\subsection{Results}
We evaluate our network variants in two different ways: (i) as models
for inferring the entire program, and (ii) as models for inferring primitives,
\ie, as object detectors.

\subsubsection{Inferring programs}
\begin{table}[t]
\begin{tabular}{l|c|c|c|c}
  Method   & IOU (k=1) & IOU (k=10) & CD (k=1) & CD (k=10) \\
  \hline
NN       & 73.9         & -          & 1.93        & -         \\
\csgnet   & 86.77     & 88.74      & 0.70     & 0.32      \\
\scsgnet & 91.33     & 93.45      & 0.60     & 0.12     
\end{tabular}
\caption{\textbf{Comparison of a NN\ baseline with the supervised network without stack (\csgnet) and
    with stack (\scsgnet)  on the synthetic 2D dataset.}
  Results are shown using Chamfer Distance (CD) and IOU metric by varying beam
  sizes ($k$) during decoding. CD is in number of pixels.}
\label{2D-dataset-table}
\end{table}

\begin{figure}[!htbp]
\centering
\includegraphics[width=\linewidth]{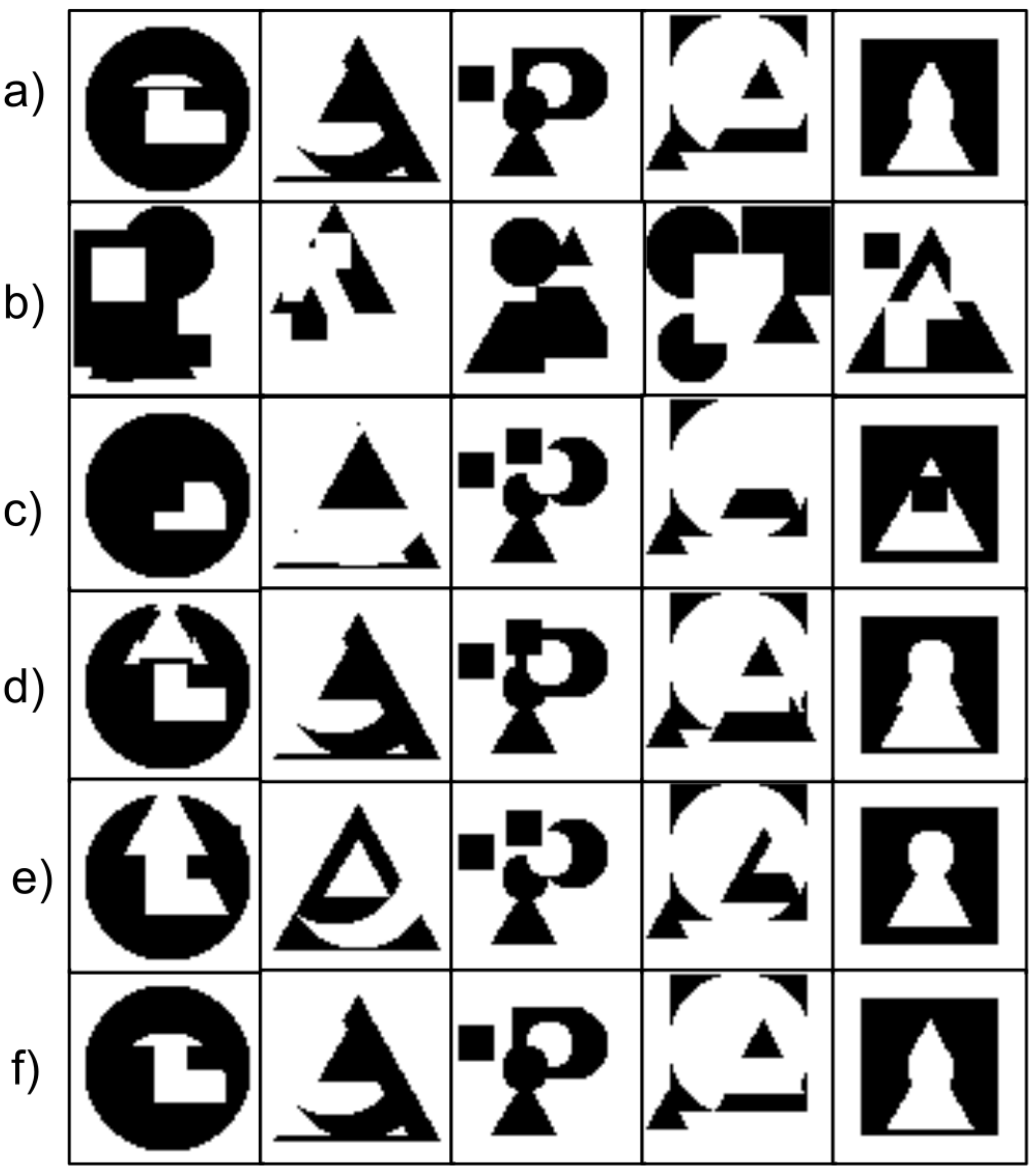}
\caption{\textbf{Comparison of performance on synthetic 2D dataset.} a) Input
  image, b) NN-retrieved image, c) top-$1$ prediction of \csgnet, d) top-$1$
  prediction of \scsgnet, e) top-$10$ prediction of \csgnet and f) top-$10$ prediction
  of \scsgnet.}
\label{fig:synth2d-comparison}
\end{figure}

\begin{table*}
\begin{tabular}{lc}
  % \setlength\tabcolsep{5.5pt}
  % \centering
  \begin{tabular}{l|l|l|cccccc}
  \multirow{2}{*}{Method} & \multirow{2}{*}{Train} & \multirow{2}{*}{Test} & \multicolumn{6}{c}{CD (@refinement iterations)}                \\\cline{4-9}
                          &                        &                       & $i$=$0$ & $i$=$1$ & $i$=$2$ & $i$=$4$ & $i$=$10$ & $i$=$\infty$ \\
  \hline
  NN                      & -                      & -                     & $1.92$    & $1.22$    & $1.13$    & $1.08$    & $1.07$     & $1.07$         \\
  \hline
  \csgnet                  & Supervised                  & k=1                   & $2.45$    & $1.2$     & $1.03$    & $0.97$    & $0.96$     & $0.96$         \\
  \csgnet                  & Supervised                  & k=10                  & $1.68$    & $0.79$    & $0.67$    & $0.63$    & $0.62$     & $0.62$         \\
  \scsgnet                & Supervised                  & k=1                   & $3.98$    & $2.66$    & $2.41$    & $2.29$    & $2.25$     & $2.25$         \\
  \scsgnet                & Supervised                  & k=10                  & $1.38$    & $0.56$    & $0.45$    & $0.40$  & $0.39$   & $0.39$       \\
  \hline
  \csgnet                  & RL                     & k=1                   & $1.40$    & $0.71$    & $0.63$    & $0.60$    & $0.60$     & $0.60$         \\
  \csgnet                  & RL                     & k=10                  & $1.19$    & $0.53$    & $0.47$    & $0.41$  & $0.41$   & $0.41$       \\
  \scsgnet                & RL                     & k=1                   & $1.27$    & $0.67$    & $0.60$    & $0.58$    & $0.57$     & $0.57$         \\
  \scsgnet                & RL                     & k=10                  & $1.02$    & $0.48$    & $0.43$    & $0.35$    & $0.34$     & $0.34$        
\end{tabular}
&
\includegraphics[width=0.4\linewidth,valign=c]{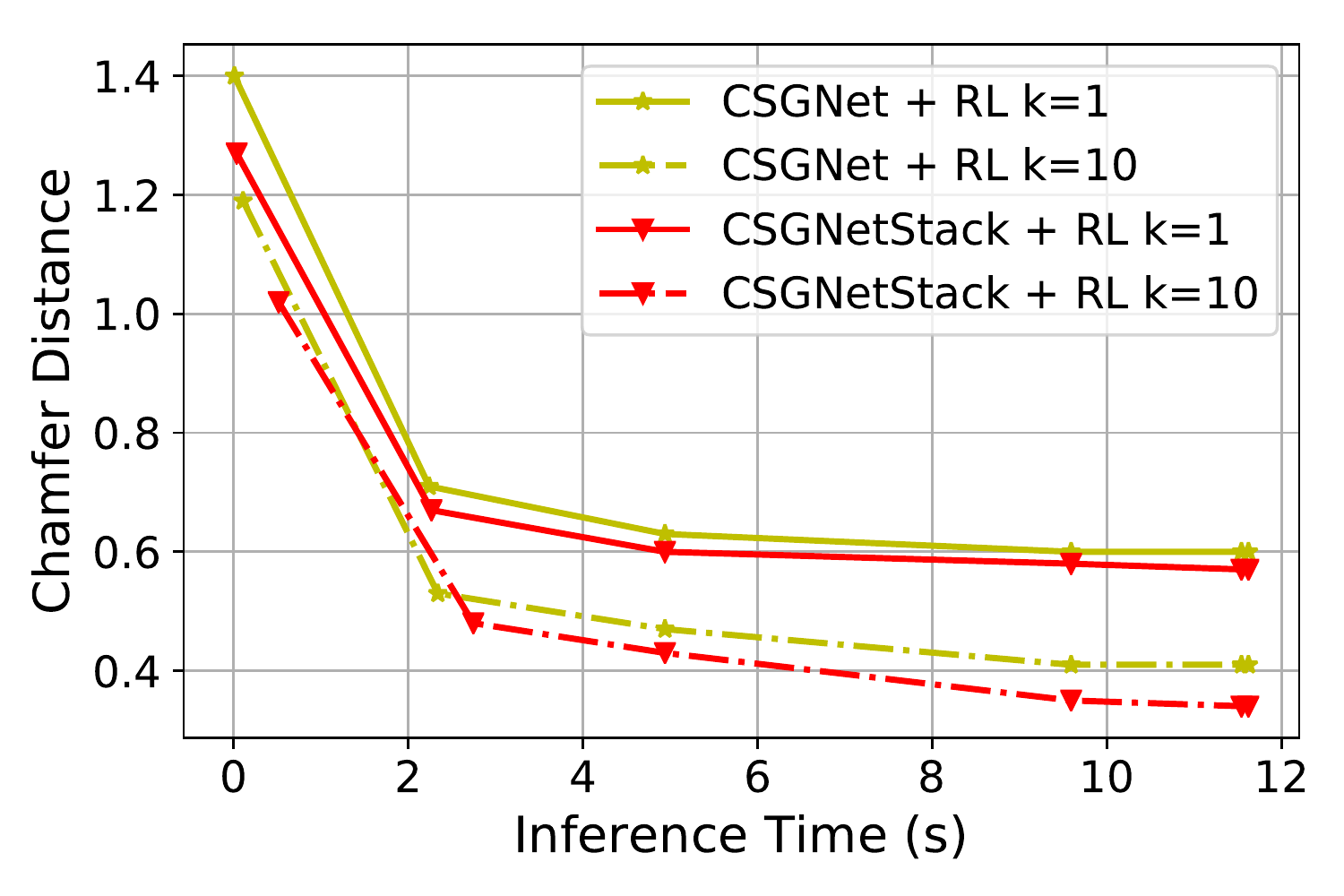}
\end{tabular}
\caption{\textbf{Comparison of various approaches on the CAD shape dataset}.
  \csgnet: neural shape parser without stack, \scsgnet: parser with stack, NN:
  nearest neighbor. Left: Results are shown with different beam sizes ($k$)
  during decoding. Fine-tuning using RL improves the performance of both
  network, with \scsgnet perfoming the best. Increasing the number of iterations
  ($i$) of visually guided refinement during testing also improves results
  significantly. $i=\infty$ corresponds to running visually guided refinement
  till convergence. Right: Inference time for different methods. Increasing
  number of iterations of visually guided refinement improves the performance,
  with least CD in a given inference time is produced by Stack based
  architecture. CD metric is in number of pixels.}
\label{table:CAD}
\end{table*}

\begin{figure*}
\centering
\includegraphics[scale=0.75]{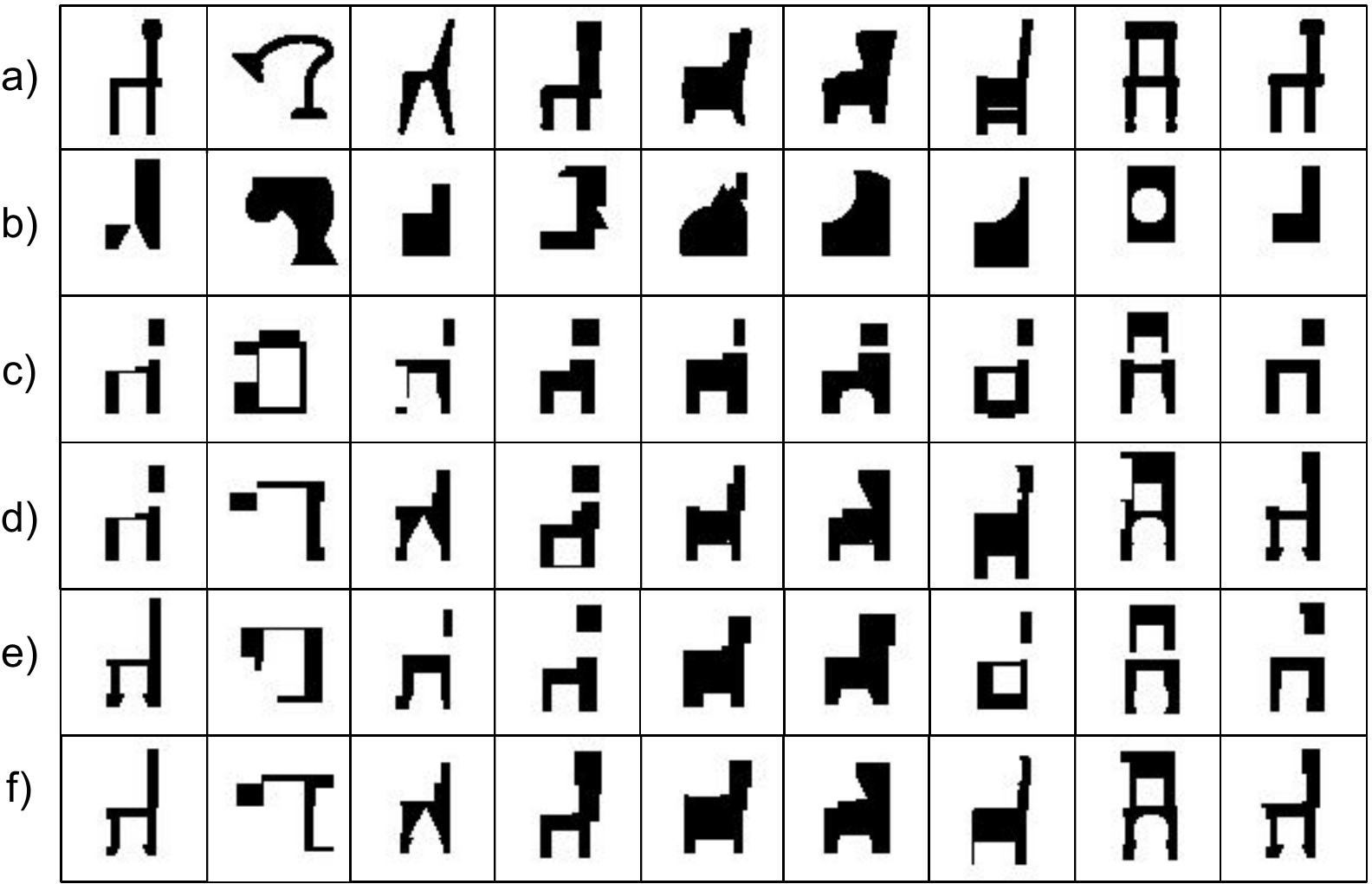}
\caption{\label{fig:cad-results}\textbf{Comparison of performance on the 2D CAD
    dataset}. a) Target image, b) NN retrieved image, c) best result from beam
  search on top of CSGNet fine-tuned with RL, d) best result from beam search on
  top of \scsgnet fine-tuned with RL, and refining results using the visually
  guided search on the best beam result of \csgnet (e) and \scsgnet (f).}
\end{figure*}

\textbf{Evaluation on the synthetic 2D shapes.} \label{synthetic-experiment}
We perform supervised learning to train our stack-based network \scsgnet and the non-stack-based network \csgnet on the training split of this synthetic dataset, and
evaluate performance on its test split under different beam sizes. We compare
with a baseline that retrieves a program in the training split using a Nearest
Neighbor (NN) approach. In NN setting, the program for a test image is retrieved
by taking the program of the train image that is most similar to the test image.

Table~\ref{2D-dataset-table} compares \scsgnet, \csgnet, and a NN\ baseline
using the Chamfer distance between the test target and predicted shapes using
the complete synthetic dataset. Our parser is able to outperform the NN method.
One would expect that NN\ would perform well here because the size of the
training set is large. However, our results indicate that our compositional
parser is better at capturing shape variability, which is still significant in
this dataset. Results are also shown with increasing beam sizes (k) during
decoding, which consistently improves performance.
Figure~\ref{fig:synth2d-comparison} also shows the programs retrieved through NN
and our generated program for a number of characteristic examples in our test
split of our synthetic dataset.

We also examine the learning capability of \scsgnet with significantly less
synthetic training dataset in comparison to \csgnet in the Figure
\ref{fig:synthetic-size}. With just $5\%$ of the total dataset, \scsgnet
performs $~80\%$ IOU ($1.3$ CD) in comparison to $70\%$ IOU ($1.7$ CD) using
\csgnet. The \scsgnet continues to perform better compared to \csgnet in
the case of more training data. This shows that incorporating the extra
knowledge in the form of an execution stack based on the proposed architecture
makes it easier to learn to parse shapes.\\

\noindent
\textbf{Evaluation on 2D CAD shapes.} \label{cad-experiment} For this
dataset, we report results on its test split under two conditions: (i) when
training our network only on synthetic data, and (ii) when training our network
on synthetic data and also fine-tuning it on the training split of rendered\ CAD
dataset using policy gradients.

Table~\ref{table:CAD} shows quantitative results on this dataset. We first
compare with the NN baseline. For any shape in this dataset, where ground truth
program is not available, NN retrieves a shape from synthetic dataset and we use
the ground truth program of the retrieved synthetic shape for comparison.

We then list the performance of \scsgnet and \csgnet trained in a supervised
manner only on our synthetic dataset. Further training with Reinforcement
Learning (RL) on the training split of the 2D\ CAD\ dataset improves the results
significantly and outperforms the NN approach by a considerable margin. This
also shows the advantage of using RL, which trains the shape parser without
ground-truth programs. The stack based network \scsgnet performs better than
\csgnet showing better generalization on the new dataset. We note that directly
training the network using RL alone does not yield good results which suggests
that the two-stage learning (supervised learning and RL) is important. Finally,
optimizing the best beam search program with visually guided refinement yielded
results with the smallest Chamfer Distance. Figure~\ref{fig:cad-results} shows a
comparison of the rendered programs for various examples in the test split of
the 2D\ CAD\ dataset for variants of our network. Visually guided refinement on
top of beam search of our two stage-learned network qualitatively produces
results that best match the input image.

\begin{figure*}
\centering
\begin{subfigure}{.5\textwidth}
  \centering
  \includegraphics[width=0.9\linewidth]{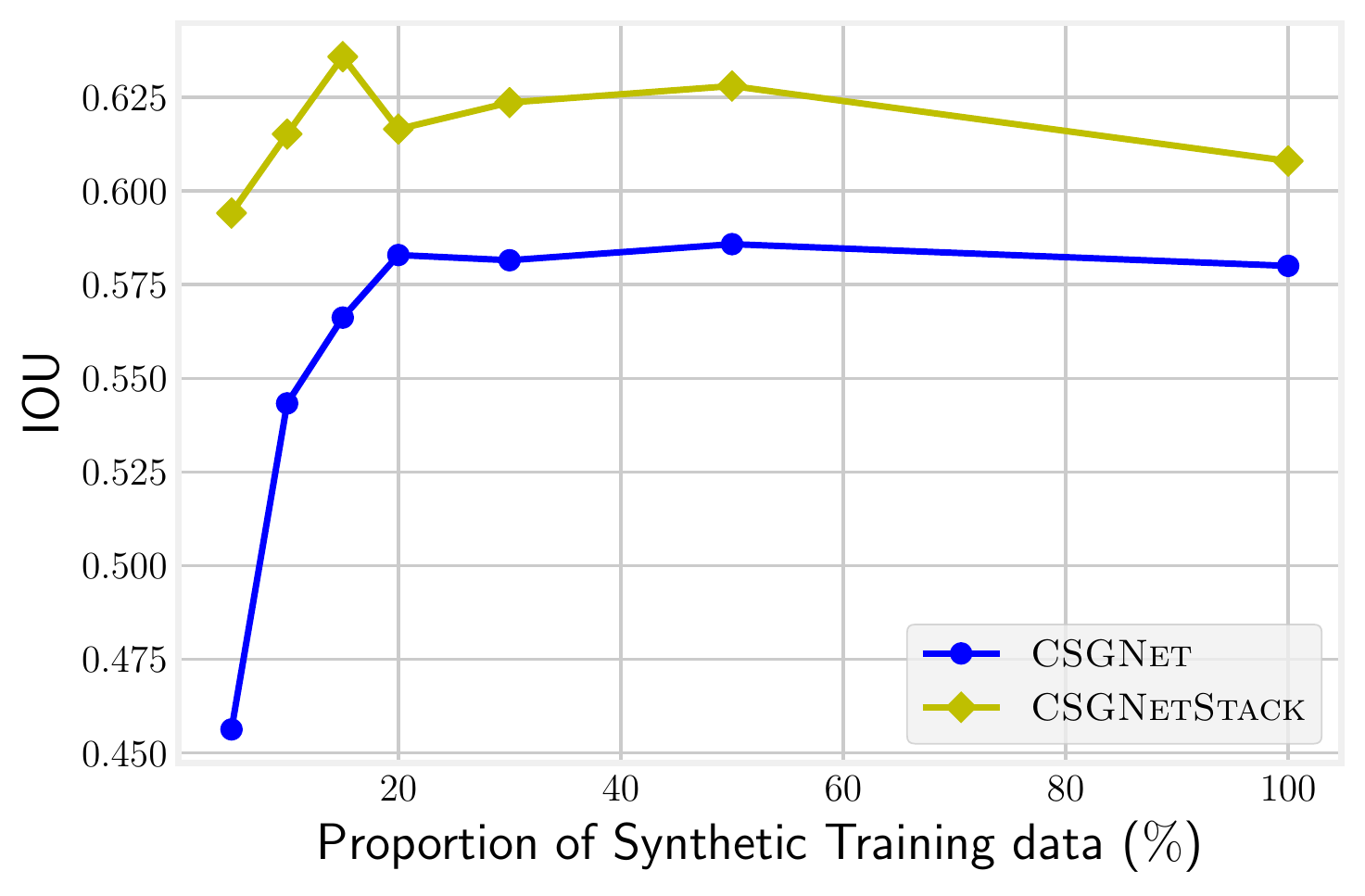}
\end{subfigure}%
\begin{subfigure}{.5\textwidth}
  \centering
  \includegraphics[width=0.9\linewidth]{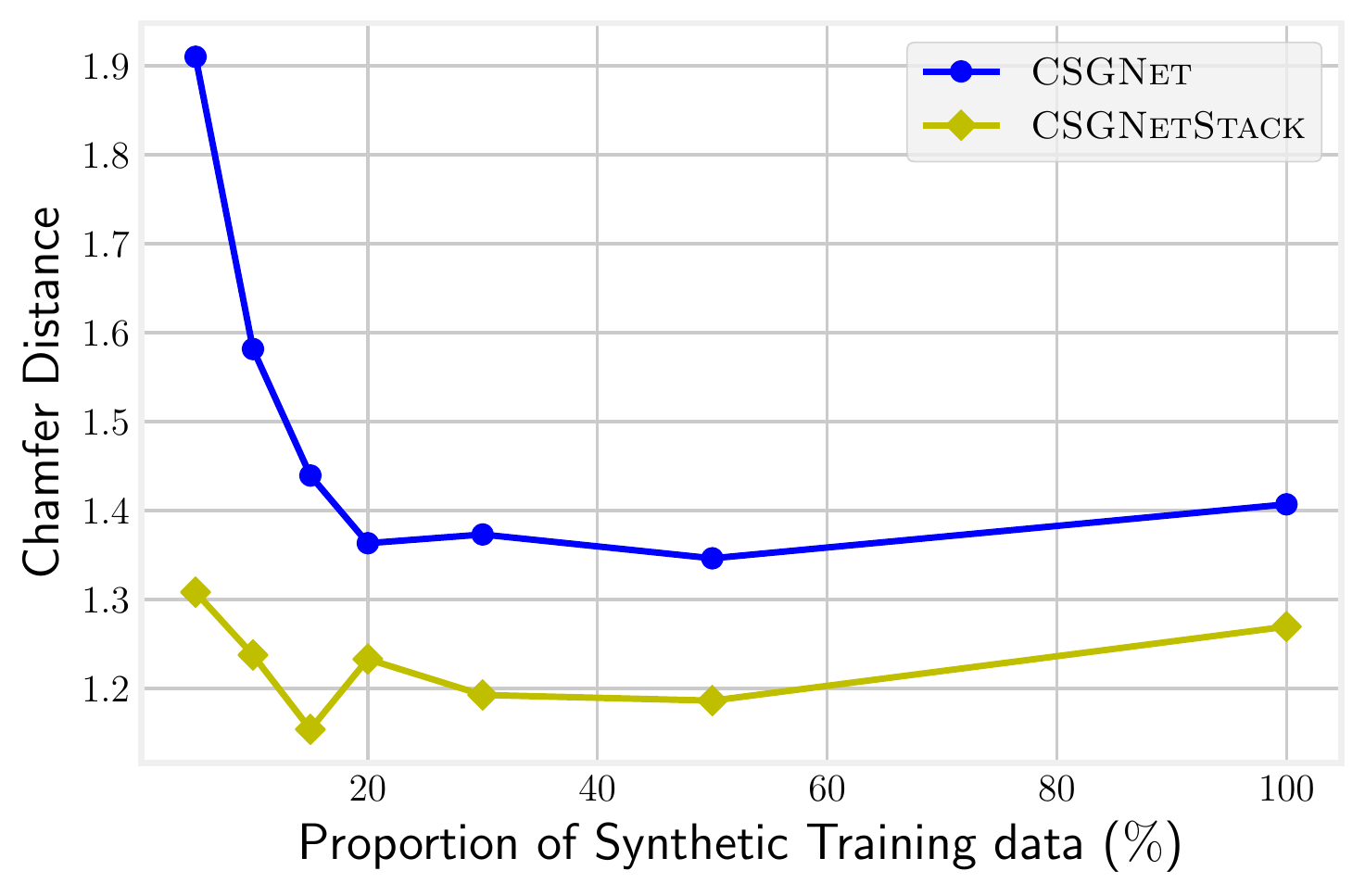}
\end{subfigure}
\caption{ \textbf{Performance (Left: IOU, Right: chamfer distance) of \csgnet and \scsgnet on
    the test split of the 2D\ CAD dataset wrt the size of the synthetic dataset used
    to pre-train the two architectures.} Pre-training is done using $x\%$ of the complete
  synthetic dataset ($x$ is shown on the horizontal axis) and fine-tuning is done on the complete CAD dataset. 
%\csgnet: our
% basic non-stack neural shape parser and \scsgnet: our variant that uses the
% execution stack. 
\scsgnet performs better while using less proportion of
the   synthetic dataset for pretraining.}
\label{fig:cad-size}
\end{figure*}

We also show an ablation study indicating how much pretraining on the synthetic
dataset is required to perform well on the CAD dataset in Figure
\ref{fig:cad-size}. With just $5\%$ of the synthetic dataset based pretraining,
\scsgnet gives $60\%$ IOU (and $1.3$ CD) in comparison to $46\%$ IOU (and 1.9
CD), which shows the\ faster learning capability of our stack based
architecture. Increasing the synthetic training size used in pretraining shows
slight decrease in performance for the \scsgnet network after $15\%$, which
hints at the overfitting of the network on the synthetic dataset domain. \\

\begin{table}[]
\begin{tabular}{l|c|c|c|clll}
\multirow{2}{*}{Method} & \multirow{2}{*}{NN} & \multicolumn{3}{c|}{3D-\csgnet}                         & \multicolumn{3}{l}{3D-\scsgnet}                                                    \\ \cline{3-8} 
                        &                     & k=$1$  & k=$5$  & \multicolumn{1}{c|}{k=$10$}          & \multicolumn{1}{l|}{k=1}  & \multicolumn{1}{l|}{k=5}  & \multicolumn{1}{l}{k=10} \\ \hline
IOU (\%)                & $73.2$              & $80.1$ & $85.3$ & \multicolumn{1}{c|}{$89.2$} & \multicolumn{1}{l|}{$81.5$} & \multicolumn{1}{l|}{$86.9$} & \multicolumn{1}{l}{$\textbf{90.5}$}
\end{tabular}
\caption{\textbf{Comparison of the supervised network (3D-$\textsc{CSGNetStack}$
    and 3D-$\textsc{CSGNet}$) 
    with NN baseline on the 3D dataset}. Results are shown using IOU(\%) metric and
  varying beam sizes ($k$) during decoding.}
\label{3D-comparison}
\end{table}
% Here, we want to evaluate the generalization capabilities of our model trained
% using synthetic dataset, on dataset for which ground truth programs are not
% available, making sure that the task can be completed using the shape
% primitives and operations that our language defines. For this, we use the 2D
% CAD dataset described in the section \ref{grammar}. Here, we create a
% \textbf{NN} baseline, that given a shape from CAD dataset retreives a nearest
% neighbors from our training split of synthetic dataset. Using reinforcement
% learning, we finetune the supervised model originally trained using synthetic
% 2D image, on this new CAD dataset. As described in the section
% \ref{inference}, we use visually guided search to finetune the parameters of
% programs produced by the network. This is mainly done to have precise
% localization and scaling, which cannot be done
% because of discrete coordinates and scaling of shape primitives.\
\begin{figure}[h]
\centering
\includegraphics[width=1.0\linewidth]{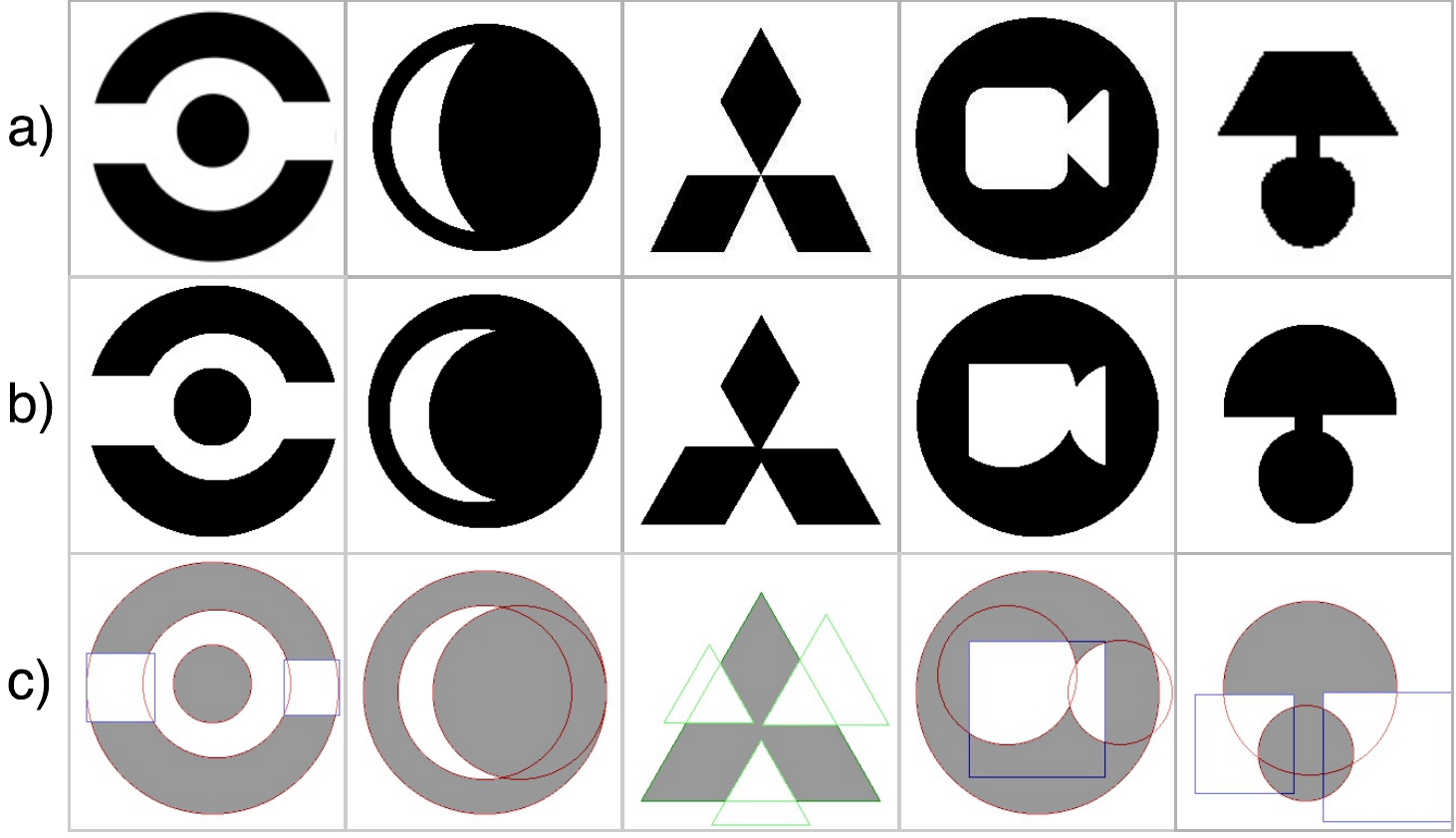}
\vskip 1mm
\caption{\textbf{Results for our logo dataset}. a) Target logos, b) output
  shapes from \csgnet and c) inferred primitives from output program. Circle
  primitives are shown with red outlines, triangles with green and squares with
  blue.}
\label{fig:logos}
\end{figure}

\noindent
\textbf{Evaluation on Logos.}
We experiment with the logo dataset described in  Section
\ref{grammar} (none of these logos participate in training). 
% For this experiment, we use our supervised model trained on synthetic 2D
% images from the section \ref{synthetic-experiment} to generate programs for
% these logos. We refine the results using search over parameters of predicted
% instructions.
Outputs of the induced programs parsing the input logos are shown in Figure
\ref{fig:logos}. In general, our method is able to parse logos into primitives
well, yet performance can degrade when long programs are required to generate
them, or when they contain shapes that are very different from our used
primitives.\\

\begin{figure}[]
\centering
\includegraphics[width=1.03\linewidth]{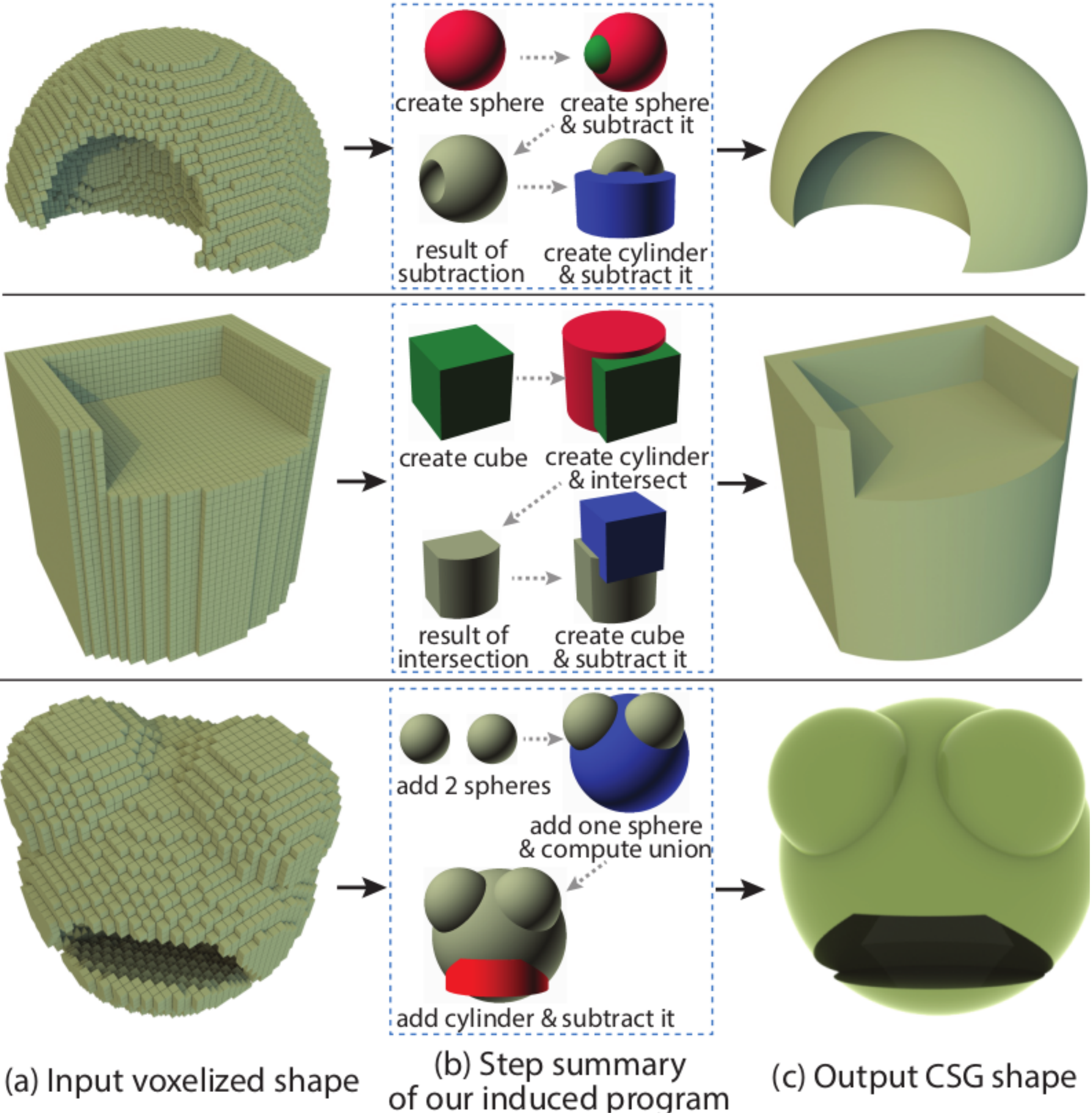}
\vskip 2mm
\caption{\textbf{Qualitative performance of 3D-\csgnet.} a) Input voxelized
  shape, b) Summarization of the steps of the program induced by 3D-\csgnet in the
  form of intermediate shapes, c) Final output created by executing induced
  program.}
\label{fig:3D-results}
\end{figure}

\noindent
\textbf{Evaluation on Synthetic 3D CSG.} Finally, we show that our approach can
be extended to 3D shapes. In the 3D\ CSG\ setting we use 3D-CSG dataset as
described in the Section \ref{grammar}. We train a stack based 3D-\scsgnet
network that takes $64\times 64 \times 64 \times (K+1)$ voxel representation of
input shape concatenated with voxel representation of stack. The input to our
3D-\csgnet are voxelized shapes in a $64$$\times$$64$$\times$$64$ grid. 
Our output is a 3D\ CSG\ program, which can be rendered as a high-resolution polygon
mesh (we emphasize that our output is not voxels, but\ CSG\ primitives and
operations that can be computed and rendered accurately). Figure
\ref{fig:3D-results} show pairs of input voxel grids and our output shapes from
the test split of the 3D\ dataset. The quantitative results are shown in the
Table~\ref{3D-comparison}, where we compare our 3D-\scsgnet and 3D-\csgnet
networks at different beam search decodings with NN method. The stack-based
network also improves the performance over the non-stack variant. The results
indicate that our method is promising in inducing correct programs for 3D
shapes, which also has the advantage of accurately reconstructing the voxelized
surfaces into high-resolution surfaces.\

\subsubsection{Primitive detection}
Successful program induction for a shape requires not only predicting correct
primitives but also correct sequences of operations to combine these primitives.
Here we evaluate the shape parser as a primitive detector (\ie, we evaluate the
output primitives of our program, not the operations themselves).
%, which is a
%simpler task). For example, the parser may detect the correct primitives but fails
%to estimate the correct order of operations resulting in a poor generated shape.
This allows us to directly compare our approach with bottom-up object detection
techniques.

In particular we compare against Faster
R-CNNs~\cite{NIPS2015-5638}, a state-of-the-art object detector.
The Faster R-CNN is based on the VGG-M
network~\cite{Chatfield14} and is trained using bounding-box and primitive
annotations based on our 2D\ synthetic training dataset. At test time the
detector produces a set of bounding boxes with associated class scores. The
models are trained and evaluated on 640$\times$640 pixel images. We also experimented
with bottom-up approaches for primitive detection based on Hough
transform~\cite{Duda} and other rule-based approaches. However, our experiments
indicated that the Faster R-CNN was considerably better.

For a fair comparison, we obtain primitive detections from \csgnet trained on the
2D\ synthetic dataset only (same as the Faster R-CNN). To obtain detection
scores, we sample $k$ programs with beam-search decoding. The primitive score is
the fraction of times it appears across all beam programs. This is a Monte Carlo
estimate of our detection score.
The accuracy can be measured through standard evaluation protocols for object
detection (similar to those in the PASCAL VOC benchmark). We report the Mean
Average Precision (MAP) for each primitive type using an overlap threshold
between the predicted and the true bounding box of $0.5$
intersection-over-union. Table~\ref{tab:primitive-detection} compares the parser
network to the Faster R-CNN approach.

Our parser clearly outperforms the Faster R-CNN detector on the squares and
triangles category. With larger beam search, we also produce slighly better
results for circle detection. Interestingly, our parser is considerably faster
than Faster R-CNN tested on the same GPU.

% \begin{figure}[!htbp] \centering
%   \includegraphics[width=0.75\linewidth]{images/todo} \label{fig:grammartree}\caption{TSNE
%     (\hl{Need to fix the script.})}\end{figure}

\begin{table}[]
\centering
\setlength{\tabcolsep}{1pt}
\begin{tabular}{l|c|c|c|c|c}
Method     & Circle     & Square     & Triangle     & Mean  & Speed (im/s)\\ \hline
Faster R-CNN   & 87.4 & 71.0 & 81.8 & 80.1 & 5\\ 
\csgnet, $k=10$ & 86.7 & 79.3 & 83.1 & 83.0 & 80\\ 
\csgnet, $k=40$ & \textbf{88.1} & \textbf{80.7} & \textbf{84.1} & \textbf{84.3} & 20
\end{tabular}
\caption{\label{tab:primitive-detection} \textbf{MAP of detectors on the
    synthetic 2D shape dataset.} We also report detection speed measured as
  images/second on a NVIDIA 1070 GPU.}
\label{proposal-comparison}
\end{table}

\section{Conclusion}
We believe that our work represents a step towards neural generation of modeling
programs given target visual content, which we believe is quite ambitious and
hard problem. We demonstrated that the model generalizes across domains,
including logos, 2D\ silhouettes, and 3D\ CAD\ shapes. It also is an effective
primitive detector in the context of 2D\ shape primitive detection.

%One direction for future investigation is to improve the parametrization of our
%output programs. Our approach produces a joint distribution over
%primitive types, sizes, locations, and modeling operations based on a single 
%categorical distribution. There can be other alternatives. For example, one possibility is to have two distinct
%categorical distributions: one for first predicting the discrete elements in the
%program like operations and primitive types, and another for primitive size and
%ocation conditioned on the first one. In our experiments, we noticed that this
%architecture does not perform better than our approach, possibly because it is
%hard to properly weigh the different categorical distributions during training.
%Another possibility is to have a categorical distribution over operations and
%primitive types, then use regression for the primitive size and location.
%However, we again observed worse performance than the presented approach. We
%surmise that selecting correct relative weights for cross entropy and regression
%losses is hard. We believe that further investigation of the output
%parametrization is a worthwhile direction, since it can affect the prediction of
%finer details in the results.
%However, due to several pooling operations in the encoder, it becomes harder to produce precise shape
%parameters. Hence, in the case of shallow encoder, producing discretized shape
%parameters on a grid and then using the post-processing refinement stage,
%produces reasonably good results.

One might argue that the 2D\ images and 3D\ shapes considered in this work are 
relatively simple in structure or geometry. However, we would like to point
out that even in this ostensibly simple application scenario (i) our method
demonstrates competitive or even better results than state-of-the-art object
detectors, and most importantly (ii) the problem of generating programs using neural networks was far
from trivial to solve: based on our experiments, a combination of memory-enabled
networks, supervised and RL strategies, along with beam and local exploration of
the state space all seemed necessary to produce good results. 

As future work, we would like to generalize our approach to longer
programs with much larger spaces of parameters in the modeling operations and
more sophisticated reward functions balancing perceptual similarity to the input
image and program length. Other promising direction is alternate strategies for combining bottom-up proposals and top-down approaches for parsing shapes, in particular, approaches based on constraint satisfaction and generic optimization.

\noindent
\textbf{Acknowledgments}. The project is supported in part by grants from the National Science Foundation (NSF) CHS-1422441,
CHS-1617333, IIS-1617917. We also acknowledge the MassTech collaborative grant for funding the
UMass GPU cluster.

\bibliographystyle{IEEEtran}
\bibliography{bibliography}

% \begin{thebibliography}{1}
% \end{thebibliography}

% \begin{IEEEbiography}{Gopal Sharma}
% \end{IEEEbiography}

% \begin{IEEEbiographynophoto}{Subhransu Maji}
% \end{IEEEbiographynophoto}

% \begin{IEEEbiographynophoto}{Evangelos Kalogerakis}
% \end{IEEEbiographynophoto}
\end{document}